\documentclass[letterpaper]{article} 
\usepackage{aaai25}  
\usepackage{times}  
\usepackage{helvet}  
\usepackage{courier}  
\usepackage[hyphens]{url}  
\usepackage{graphicx} 
\usepackage{natbib}  
\usepackage{caption} 
\usepackage{algorithm}
\usepackage{algorithmic}

\usepackage{newfloat}
\usepackage{listings}
\usepackage{newfloat}
\usepackage{listings}
\usepackage{bm}
\usepackage{amsmath}
\usepackage{amsthm}
\usepackage{color}

\usepackage{amsthm,amsmath,amssymb}

\newtheorem{thm}{Theorem}

\newtheorem{cor}{Corollary}
\newtheorem{remark}{Remark}

\DeclareCaptionStyle{ruled}{labelfont=normalfont,labelsep=colon,strut=off} 
\floatstyle{ruled}
\newfloat{listing}{tb}{lst}{}
\floatname{listing}{Listing}
\title{Conditional Diffusion Models Based Conditional Independence Testing}
\author{
 Yanfeng Yang\textsuperscript{\rm 1}\equalcontrib,
 Shuai Li\textsuperscript{\rm 1}\equalcontrib, Yingjie Zhang\textsuperscript{\rm 1}\equalcontrib, Zhuoran Sun\textsuperscript{\rm 1},  Hai Shu\textsuperscript{\rm 2},  
 Ziqi Chen\textsuperscript{\rm 1}\thanks{Corresponding author: Ziqi Chen (zqchen@fem.ecnu.edu.cn, chenzq453@163.com).},   Renming Zhang\textsuperscript{\rm 3}
}
\affiliations{
    \textsuperscript{\rm 1}School of Statistics, KLATASDS-MOE, East China Normal University, Shanghai, China\\
 \textsuperscript{\rm 2}  Department of Biostatistics,
School of Global Public Health,
New York University, New York, USA\\
\textsuperscript{\rm 3} Department of Computer Science, Boston University,  Boston,  USA

}

\usepackage{bibentry}

\begin{document}

\maketitle

\begin{abstract}
Conditional independence (CI)  testing is a fundamental task in modern statistics and machine learning. The conditional randomization test (CRT) was recently introduced to test whether two random variables, $X$ and $Y$, are conditionally independent given a potentially high-dimensional set of random variables, $Z$. The CRT operates exceptionally well under the assumption that the conditional distribution $X|Z$ is known. However, since this distribution is typically unknown in practice, accurately approximating it becomes crucial.
In this paper, we propose using  conditional diffusion models (CDMs) to learn the distribution of $X|Z$. Theoretically and
empirically, it is shown that CDMs  closely approximate the true conditional distribution. Furthermore, CDMs offer a more accurate approximation of 
$X|Z$ compared to GANs, potentially leading to a CRT that performs better than those based on GANs.
To accommodate complex dependency structures, we utilize a computationally efficient classifier-based conditional mutual information (CMI) estimator as our test statistic. The proposed testing procedure performs effectively without requiring assumptions about specific distribution forms or feature dependencies, 
and is capable of handling mixed-type conditioning sets that include both continuous and discrete variables. Theoretical analysis  shows that our proposed test achieves a valid control of the type I error.
A series of experiments on synthetic data 
demonstrates that our new test 
effectively controls both type-I and type-II errors,  even in  high dimensional scenarios.


\end{abstract}

%

\section{Introduction}
\label{Intro}
Conditional independence (CI) is an important concept in statistics and machine learning. Testing conditional independence plays a central role in classical problems such as causal inference \cite{pearl1988probabilistic,spirtes2000causation}, graphical models \cite{lauritzen1996graphical,koller2009probabilistic}, and variable selection \cite{dai2022significance}. It is widely used in various scientific problems, including gene regulatory network inference \cite{dai2024gene} and personalized therapies \cite{khera2017genetics,zhu2018causal}. 
We consider testing whether two random variables, 
$X$ and 
$Y$, are independent given a random vector 
$Z$, based on observations of the joint density $p_{X,Y,Z}(x,y,z)$. Specifically, we test the following hypotheses:
\begin{equation*}
H_0: X \perp \!\!\! \perp Y|Z ~~\mbox{versus}~~ H_1:X \not \! \perp \!\!\! \perp Y|Z,
\end{equation*}
where $\perp \!\!\! \perp$ denotes the independence. In practical genome-wide association studies, $X$ represents a specific genetic variant, $Y$ denotes disease status, and $Z$ accounts for the rest of the genome. By conditioning on $Z$, we can evaluate whether the genetic variant $X$ has an effect on the disease status $Y$ \cite{liu2022fast} by CI testing.  CI testing becomes particularly challenging due to the high-dimensionality of the conditioning vector $Z$ \cite{bellot2019conditional,shi2021double}. Moerever, the presence of mixed discrete and continuous variables in $Z$
as in many real-world applications  presents further challenges for the testing \cite{mesner2020conditional,zan2022conditional}. 


Recently, there has been a large and growing literature on CI testing, and for a more comprehensive review, we refer readers to \citeauthor{li2020nonparametric} \shortcite{li2020nonparametric}. The metric-based tests (e.g., \citeauthor{su2008nonparametric} \shortcite{su2008nonparametric}, \citeauthor{su2014testing} \shortcite{su2014testing}, \citeauthor{wang2015conditional} \shortcite{wang2015conditional}) employ some kernel smoothers to estimate the conditional characteristic function or the distribution function of $Y$ given $X$ and $Z$. However, due to the curse of dimensionality,  these tests are usually not suitable when the conditioning vector $Z$ is high-dimensional. The kernel-based tests, such as \citet{fukumizu2007kernel}, \citet{zhang2011kernel}, and \citet{scetbon2022asymptotic}, represent probability distributions as
elements of a reproducing kernel Hilbert space (RKHS), which enables us to understand properties of these distributions using Hilbert space operations. However, these tests based on asymptotic distributions may exhibit inflated type-I errors or inadequate power when dealing with 
high-dimensional $Z$ \cite{doran2014permutation,runge2018conditional,shi2021double}. The most relevant work to ours is the conditional randomization test (CRT) proposed by \citeauthor{candes2018panning} \shortcite{candes2018panning}, which assumes the true conditional distribution of $X$ given $Z$, denoted by $P(\cdot|Z)$, is known. It is theoretically proven that the CRT maintains validity by ensuring that the type I error does not exceed the significance level $\alpha$ \cite{berrett2020conditional,liu2022fast}.


However, the true conditional distribution $P(\cdot|Z)$ is rarely known in practice, several methods have been developed to approximate the $P(\cdot|Z)$. The smoothing-based methods \cite{hall2004cross,hall2005approximating,izbicki2017converting} suffer from the curse of dimensionality, and their performance deteriorates sharply when the dimensionality of $Z$ becomes large. \citeauthor{bellot2019conditional} \shortcite{bellot2019conditional}  developed a  Generative Conditional Independence Test (GCIT) by using Wasserstein generative adversarial networks \citep[WGANs;][]{arjovsky2017wasserstein} to approximate $P(\cdot|Z)$. \citeauthor{shi2021double} \shortcite{shi2021double} proposed to use the Sinkhorn GANs \cite{genevay2018learning} to approximate $P(\cdot|Z)$. However, the training of GANs is often unstable, with the risk of collapse if hyperparameters and regularizers are not carefully chosen \cite{dhariwal2021diffusion}. 
The potentially inaccurate learning of conditional distributions using GANs can lead to inflated type I errors in GANs-based CI tests \cite{li2023nearest}.
\citeauthor{li2023nearest} \shortcite{li2023nearest} introduced a method using the 1-nearest neighbor technique to generate samples from the approximated conditional distribution of 
$X$ given 
$Z$. However, their approach necessitates dividing the dataset into two segments. Consequently, only one-third of the total samples are allocated to the testing dataset used for calculating test statistics, which reduces the test's statistical power.   \citeauthor{li2024k} \shortcite{li2024k} proposed utilizing a k-nearest-neighbor local sampling strategy to generate samples from the approximated conditional distribution of $X$ given $Z$. Nevertheless, this approach encounters issues with insufficient sample diversity, resulting in unstable performance of the CI test, particularly when the conditioning variables $Z$ include both continuous and discrete  variables.

Diffusion models, which have recently emerged as a notable class of generative models, have attracted significant attention \cite{ho2020denoising,song2021score,yang2023diffusion}. Unlike GANs, diffusion models provide a much more stable training process and generate more realistic samples \cite{dhariwal2021diffusion,songdenoising}. They have achieved great success in various tasks, such as image generation \cite{ho2020denoising} and video generation \cite{ho2022video}. In this paper, we propose using conditional diffusion models to approximate $P(\cdot | Z)$ and  generating samples from the approximated conditional distribution. Moreover, as highlighted in \citeauthor{li2023nearest} \shortcite{li2023nearest}, the choice of test statistics in CRT procedure is crucial for achieving  adequate statistical power as well as controlling type I errors. Conditional mutual information (CMI) for $(X, Y, Z)$, denoted as $I(X;Y|Z)$, provides a strong theoretical guarantee for conditional dependence relations. Specifically,  
$I(X;Y|Z)=0\iff X \perp \!\!\! \perp Y|Z$ \cite{cover2012elements}. 
In this paper, we adopt the classifier-based CMI estimator as the test statistic 
\cite{mukherjee2020ccmi,li2024k}. 

Our main contributions  are summarized as follows. First, we  propose, for the first time, using conditional diffusion models to generate samples for the conditional distribution in the CI testing task. 
Compared to GANs, this method is not only more stable during training but also demonstrates significant advantages in sample quality and diversity. It is theoretically and empirically shown that the distribution of the generated samples is very close to the true conditional distribution. 
Second, we use a computationally efficient classifier-based
CMI estimator as the test statistic, which  captures intricate
dependence structures among variables.
Third, theoretical analysis demonstrates that our proposed test achieves a valid control of the type I error asymptotically. 
Fourth, 
our empirical evidence demonstrates that, compared to state-of-the-art methods, our test effectively controls type I error while maintaining sufficient power under $H_1$. This remains true even when handling high-dimensional data and/or mixed-type conditioning sets that include both continuous and discrete variables.


\section{The Proposed  Approach}
\subsection{The Conditional Randomization Test (CRT)}
Our work builds on the conditional randomization test (CRT) proposed by \citeauthor{candes2018panning} \shortcite{candes2018panning}, which, however, assumes that the true conditional distribution $P(\cdot|Z)$ is known. Specifically, consider
$n$  i.i.d.  copies $\mathcal{D}_{\text{T}}=\{(X_i,Y_i,Z_i): i=1, \ldots, n\} $ of $(X, Y, Z)$.   If $P(\cdot|Z)$ is known, conditionally on $\bm{Z}=(Z_1, \ldots, Z_n)^T$, one can independently draw $X_{i}^{(b)}\sim P(\cdot|Z_i)$ for each $i$ across $b=1, \ldots, B$ such that  all  $\bm{X}^{(b)}:=(X_{1}^{(b)}, \ldots, X_{n}^{(b)})^T$ are independent of $\bm{X}:=(X_1, \ldots, X_n)^T$ and $\bm{Y}:=(Y_1, \ldots ,Y_n)^T$, where $B$ is the number of repetitions. Thus, under the null hypothesis $H_0:X\perp \!\!\! \perp Y|Z$, we have $(\bm{X}^{(b)}, \bm{Y}, \bm{Z})\overset{d}{=}(\bm{X}, \bm{Y}, \bm{Z})$ for all $b$, where $\overset{d}{=}$ denotes equality in distribution. A large difference between $(\bm{X}^{(b)}, \bm{Y}, \bm{Z})$ and $(\bm{X}, \bm{Y}, \bm{Z})$ can be regarded as a strong evidence against $H_0$.  Statistically, for any  test statistic $\bm{T}(\bm X,\bm Y,\bm Z)$, we can  calculate the  $p$-value of the CI test by
\begin{equation}\label{CRT}
    \frac{1+\sum_{b=1}^{B}\bm{\mbox{I}}\{\bm{T}(\bm X^{(b)},\bm Y,\bm Z)\geq \bm{T}(\bm X,\bm Y,\bm Z)\}} {1+B},
\end{equation}
where $\bm{\mbox{I}}\{ \cdot\}$ is the indicator function. Under the null hypothesis $H_0$, since the ($B+1$) triples $(\bm{X}, \bm{Y}, \bm{Z}),(\bm{X}^{(1)}, \bm{Y}, \bm{Z}),\ldots,(\bm{X}^{(B)}, \bm{Y}, \bm{Z})$ are exchangeable, the above $p$-value is valid. Specifically,   $P(p\leq \alpha |H_{0})\leq \alpha $ holds for any $\alpha \in (0,1)$ \cite{candes2018panning,berrett2020conditional}.



\subsection{Methodology for Sampling}
The traditional CRT procedure assumes that the true conditional distribution $P(\cdot|Z)$ is known.  However, in practice, this distribution is seldomly known.
We propose using  score-based conditional diffusion models to approximate the conditional distribution of $X$ given $Z$. 
Specifically, we have an unlabelled data set $\mathcal{D}_{\text{U}}$ that consists of $N$ i.i.d. samples $\{ (X_i^{\text{U}},Z_i^{\text{U}}): i=1,\ldots,N \}$ 
from the distribution $P_{X,Z}$. We aim to accurately recover the true distribution of $X$ conditioning on $Z$ using $\mathcal{D}_{\text{U}}$. 


The score-based  conditional diffusion models involve two stochastic processes: the forward process and the reverse process. Specifically, let $t \in [0,T]$  be the time index in forward process and reverse process, and denote $X:=X(0) \in \mathbb{R}^{d_x}$. 
Conditioned on $Z$, the forward process is presented as:
\begin{equation*}
\label{forward process}
dX(t)=-\dfrac{1}{2}X(t)dt+dB(t), 
\end{equation*}
where $X(0) \sim P(\cdot|Z)$ and $B(t)$ is a standard Brownian motion in $\mathbb{R}^{d_x}$. At any time $t$, let $p_t(\cdot|Z)$ and $P_{t}(\cdot|Z)$ be the conditional density and  distribution of $X(t)|Z$, respectively. By the property of Ornstein Uhlenbeck (OU) process, we  derive 
\begin{equation}
\label{expression of x0}
    X(t) \overset{d}{=} \exp(-t/2) X(0)+ \sqrt{1-\exp(-t)}\epsilon, 
\end{equation} 
where $\epsilon \sim N(0,I_{d_x})$ and $I_{d_x}$ is the $d_x$-dimensional identity matrix.
It can be deduced that as $T$ approaches infinity, $X(T) \sim N(0,I_{d_x})$. In practice, however, $ X(t) $ is stopped at a sufficiently large $ T $ to ensure computational feasibility.

According to \citeauthor{song2021score} (\citeyear{song2021score}), the true reverse process is:
\begin{equation*}
\label{reverse process}
\begin{array}{c}
    d\overline{X}(t)=\left[ \dfrac{1}{2}\overline{X}(t) + \nabla \log p_{T-t}(\overline{X}(t)|Z) \right] dt +d\overleftarrow{B}(t), \\ 
    \overline{X}(0) \sim P_{T}(\cdot|Z),
\end{array}
\end{equation*}
where $\overleftarrow{B}(t)$ is a standard Brownian motion in reversed process. We observe that the distributions of $\overline{X}(t)$ and $X(T-t)$ are identical. Thus,   
the  conditional density and distribution of $\overline{X}(t)| Z$ are  $p_{T-t}(\cdot|Z)$ and $P_{T-t}(\cdot|Z)$, respectively. 
 $\nabla \log p_{T-t}(\cdot|z)$ is the score function of $\overline{X}(t)$ conditioned on $Z=z$. 
The analytical solution for the conditional score function is unattainable. Here, we utilize a ReLU neural network $\widehat{s}(x,z,t) \in \mathcal{F}$ to approximate it, where $\mathcal{F}$ is the function class of ReLU neural networks. Following  \citeauthor{ho2020denoising} \shortcite{ho2020denoising} and \citeauthor{song2021score} \shortcite{song2021score}, we train $\widehat{s}(x, z, t)$ by approximating the conditional score function through minimization of the following objective:
\begin{equation*}
\label{score matching}
    \mathbb{E}_t\mathbb{E}_{X(0),Z}\mathbb{E}_{X(t)} \| \widehat{s}(X(t),Z,t)-\nabla \log \phi (X(t),X(0)) \|_2^2, 
\end{equation*}
where
\begin{equation*}
\begin{array}{c}
    X(t) \sim N(\exp{(-t/2)}X(0),~ [1-\exp{(-t)]}I_{d_x}), \\
    \phi (X(t),X(0))=\exp\left[ \dfrac{ \| X(t)-\exp{(-t/2)} X(0)\|^2}{-2 \left( 1-\exp{(-t)} \right)}\right],\\
    (X(0),Z) \sim P_{X,Z}, ~ t \sim \text{Uniform} [t_{\min},T], 
\end{array}
\end{equation*}
and $t_{\min}$ is an early-stopping time close to zero ensuring the denominator in $\phi$ is not zero. 
As described in (\ref{expression of x0}), when $T$ is sufficiently large, $P_{T}(\cdot|Z)$ can be approximated by $N(0,I_{d_x})$. Therefore, given $Z$, we propose the following reverse process with the approximated score function to generate pseudo samples:
\begin{equation}
\label{reverse process with N(0,1)}
\begin{array}{c}
    d\overleftarrow{X}(t)=\left[ \dfrac{1}{2}\overleftarrow{X}(t) + \widehat{s}(\overleftarrow{X}(t),Z,T-t) \right] dt +d\overleftarrow{B}(t), \\ 
    \overleftarrow{X}(0) \sim N(0,I_{d_x}).
\end{array}
\end{equation}
Let \(\overleftarrow{X}(T - t_{\min})\) be the pseudo-sample generated by (\ref{reverse process with N(0,1)}). As demonstrated in Theorem \ref{thm1}, this pseudo-sample has a conditional distribution \(\widehat{P}(\cdot|Z)\) that approximates the true conditional distribution \(P(\cdot|Z)\) effectively \cite{Fu2024UnveilCD}. 
Algorithm \ref{algo:train card}  outlines the training procedure for the conditional score matching models, whereas the sampling process from the reverse process is detailed in Algorithm \ref{algo:card sampler}.

\begin{algorithm}[htb]
\caption{Training the conditional score matching models}  
\label{algo:train card}  
\textbf{Input}: A data set with $N$ i.i.d. samples $\{ X_i^{\text{U}},Z_i^{\text{U}} \}_{i=1}^{N}$. \\
\textbf{Output}: The score network $\widehat{s}$. 
\begin{algorithmic}[1]
    \STATE Let $\{ X_i (0) \}_{i=1}^N = \{ X_i^{\text{U}} \}_{i=1}^{N}$
    \STATE Initialize a deep neural network $\widehat{s}(x,z,t)$
    \WHILE{not converge}
        \STATE Draw $t \sim \text{Uniform}[t_{\min},T]$
        \STATE Draw $\epsilon_1,\ldots,\epsilon_N \sim N(0,I_{d_x})$~\\
        \STATE Let $X_i(t) = \exp(-t/2) X_i(0)+ \sqrt{1-\exp(-t)}\epsilon_i$
        \STATE Compute $\textit{{L}}_{\text{score}}=\sum_{i=1}^N \| \widehat{s}(X_i(t),Z_i^{\text{U}},t)+ [X_i(t)-\exp(-t/2)X_i(0)]/(1-\exp(-t))\|_2^2$
        \STATE Take optimization step on $\nabla\textit{{L}}_{\text{score}}$ and update the parameters of $\widehat{s}$
    \ENDWHILE
    \STATE \textbf{Return} $\widehat{s}$ 
\end{algorithmic}
\end{algorithm}

\begin{algorithm}[htb]
\caption{Sampling from  score-based conditional diffusion models 
}  
\label{algo:card sampler}  
\textbf{Input}: A sample $Z \sim P_Z$, the score network $\widehat{s}$, and sample step $K$.\\
\textbf{Output}: Pseudo sample $\widehat{X}$. 
\begin{algorithmic}[1]
    \STATE Evenly divide $[0,T-t_{\min}]$ into $t_0=0<t_1< \ldots <t_K=T-t_{\min}$ and let $\Delta t= (T-t_{\min})/K$
    \STATE Draw $\overleftarrow{X}(0) \sim N(0,I_{d_x})$.
    \FOR{$k=0$ to $K-1$}
        \STATE Draw $\epsilon_k \sim N(0,I_{d_x})$
        \STATE Let $\overleftarrow{X}(k+1)=\overleftarrow{X}(k)+\sqrt{\Delta t}\epsilon_k $
        \STATE  $ \quad \quad \quad \quad \quad \quad  +\left[ \dfrac{1}{2}\overleftarrow{X}(k) + \widehat{s}(\overleftarrow{X}(k),Z,T-t_{k}) \right]\Delta t$
    \ENDFOR
    \STATE Let $\widehat{X}=\overleftarrow{X}(K)$
    \STATE \textbf{Return} $\widehat{X}$
\end{algorithmic}  
\end{algorithm}


\subsection{Theoretical Guarantee for Sampling Quality}
\label{tgsq}
We present a theoretical result showing that the distribution of the generated samples closely resembles the true conditional distribution.
Specifically,
denote the true conditional density of $X$ given $Z$ as $p(\cdot|Z)$, and the density of $\widehat{X}$ sampled from Algorithm \ref{algo:card sampler} given $Z$ as $\widehat{p}(\cdot|Z)$.
The total variation distance between two density functions $p_1$ and $p_2$ is defined as $d_{\text{TV}}(p_1,p_2)=\frac{1}{2}\int|p_1(x)-p_2(x)|dx$. 
Define 
\begin{align*}
    &\Gamma_1(k,\alpha)=\frac{k+\alpha}{d_x+d_z+2k+2\alpha}, \\
    &\Gamma_2(k,\alpha)=\max\left( \frac{19}{2},\frac{k+\alpha+2}{2} \right),
\end{align*} where $k$ and $\alpha$ are 
defined in Supplementary Materials. The proof of Theorem \ref{thm1} is deferred to Supplementary Materials.

\begin{thm} 
\label{thm1}
Under Assumptions 1 and 2 in the Supplementary Material, taking early stopping time $t_{\min}=N^ {-4\Gamma_1(k,\alpha)-1}$ and terminal time $T=2\Gamma_1(k,\alpha)\log N$, when $N\to \infty$, we have 
\end{thm}
\begin{equation*}
\begin{array}{l}
    \mathbb{E}_{\{X_i^{\text{U}},Z_i^{\text{U}}\}_{i=1}^{N}}\mathbb{E}_Z[d_{\text{TV}}(p(\cdot|Z),\widehat{p}(\cdot|Z))] \\ \\
    =O\left( N^ {-\Gamma_1(k,\alpha)} \cdot \left(\log N \right) ^{\Gamma_2(k,\alpha)} \right). 
\end{array}
\end{equation*}


\subsection{Empirical Evidence for Sampling Quality}
In this subsection, we investigate the empirical performance of our proposed method for approximating the conditional distribution in sample generation in the CRT framework. Specifically, we  compare it with several well-known existing conditional distribution estimating methods in CI test, including WGANs \cite{bellot2019conditional}, Sinkhorn GANs \cite{shi2021double}, and k-nearest neighbors (k-NN) \cite{li2024k}. 
We consider 
the following three models.
\begin{itemize}

\item Model M1: 
$X=Z_1^2+\exp(Z_2+Z_3/3)+Z_4-Z_5+0.5 (1+Z_2^2+Z_5^2) \epsilon,$ where $ Z_1,\ldots,Z_5,\epsilon$ $\mathop{\sim}\limits_{}^{i.i.d.}$ $N(0,1)$.

\item Model M2: 
$X=(5+Z_1^2/3+Z_2^2+Z_3^2+Z_4^2+Z_5^2) \exp(r),$ where
$r\sim B\times N(-2,1)+(1-B)\times N(2,1)$,
$B\sim \text{Bernoulli}(1,0.5)$, and $Z_1,\ldots,Z_5$ $\mathop{\sim}\limits_{}^{i.i.d.}$ $N(0,1)$. 

\item Model M3: 
$X=\sum_{i=1}^{13} Z_i/13+0.33 \epsilon$, where $Z_1,\ldots,Z_{10}, \epsilon  \mathop{\sim}\limits_{}^{i.i.d.}  N(0,1),$ and $Z_{11},\ldots,Z_{20} \mathop{\sim}\limits_{}^{i.i.d.} 2 \cdot \text{Bernoulli}(1,0.5)-1$. 
\end{itemize}

Models M1 and M2 exhibit complex, nonlinear and non-monotonic relationships between $X$ and $Z$. Model M3 involves a mixed-type 
$Z$ consisting of both continuous and discrete variables, 
where we model the true relationship between $X$ and $Z = (Z_1, \ldots, Z_{20})$ using only $(Z_1, \ldots, Z_{13})$.
 For each model, we use 500 samples to learn the conditional distribution for each method.

\begin{figure}[ht]
    \centering
    \begin{minipage}[t]{0.83\linewidth}
        \centering
        \includegraphics[width=1\linewidth]{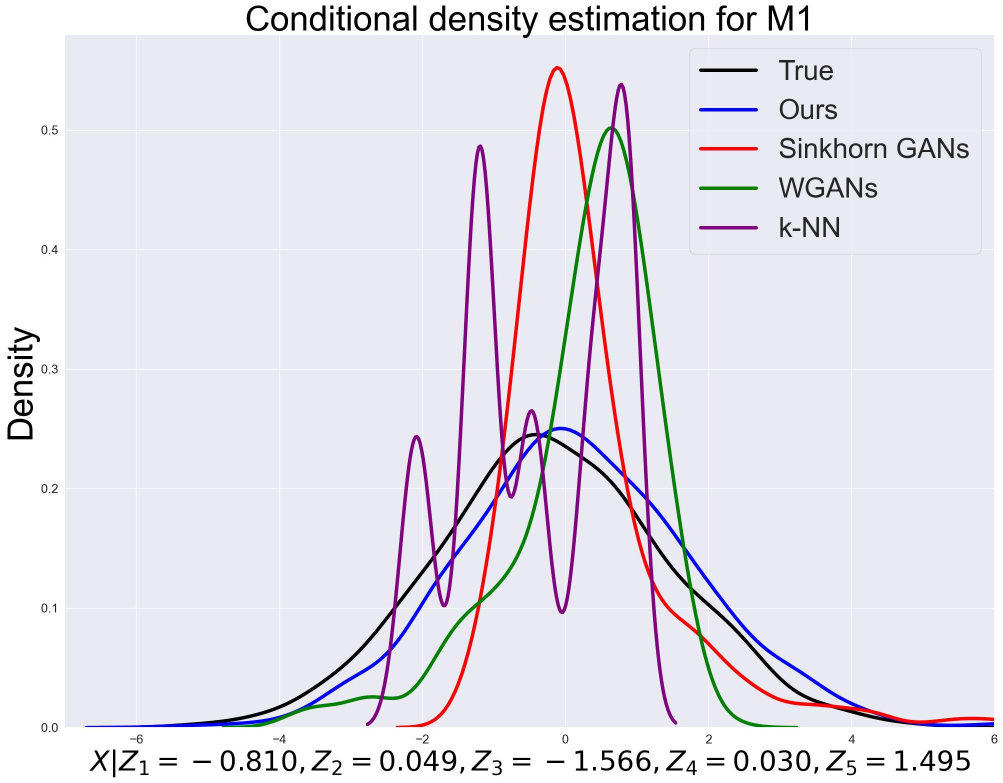}
    \end{minipage}
    
    \caption{Comparison of conditional \mbox{density} 
    estimators on Model M1. $Z=(-0.810, 0.049, -1.566, 0.030, 1.495)$.}
    \label{fig:M1 to M4}
 \end{figure}

For each model,
we estimate the conditional density functions using the 500 samples generated from each 
method with kernel smoothing \cite{kde_plot}. Figure \ref{fig:M1 to M4} and Figure \ref{fig:M2 to M3} (Supplementary Material) display the estimated conditional density functions for a randomly generated value of $Z$. The results demonstrate that our conditional diffusion estimation method yields better conditional density estimators than WGANs, Sinkhorn GANs and k-NN. Further, it can be observed that the samples generated by k-NN lack diversity.

To further evaluate these methods, we compute the mean squared errors between the quantiles of  the generated samples and those of the true conditional distribution. 
Specifically, a value \( z \) is randomly sampled. Given \( Z = z \), we generate 500 samples for each conditional distribution estimating method. We then calculate the squared error between the \(\tau\) quantile of the generated samples and the \(\tau\) quantile of the true distribution over $\tau \in \{0.05, 0.25, 0.50, 0.75,  0.95\}$. This procedure is repeated 100 times, and the mean squared errors for each $\tau$ are reported in Table \ref{quantile mse}.
Our proposed sampling method performs best on Models M1 and M3. It is also quite competitive on Model M2, though it is not always the  top performer. 

\begin{table*}[ht]
\centering
\begin{tabular}{cccccc}
\hline
Model &Quantile & Ours & WGANs & Sinkhorn GANs & k-NN\\

\hline
&0.05 & \textbf{1.731(3.740)} & 7.478(6.372) & 5.474(5.910)  &  2.062(5.962)\\
&0.25 & \textbf{0.508(0.998)} & 3.676(4.585) & 2.384(2.468)  &  1.201(2.193)\\
M1&0.50 & \textbf{0.405(0.421)} & 2.568(3.408) & 2.595(2.905)&  0.876(1.773)\\
&0.75 & \textbf{0.951(0.936)} & 2.364(2.509) & 4.451(4.221)  &  1.601(2.278)\\
&0.95 & \textbf{3.758}(4.589) & 3.871\textbf{(1.386)} & 9.716(9.708)  &  4.617(6.932)\\
\hline

&0.05 & 2.684(4.288) & 4.465\textbf{(1.684)} & 7.733(6.258)   &   \textbf{1.014}(3.343)\\
&0.25 & \textbf{3.609(3.985)} & 6.929(5.917) & 11.670(11.963)      &   8.640(24.409)\\
M2&0.50 & \textbf{9.072}(11.561) & 18.820(13.056) & 25.303(13.502)   &   36.565\textbf{(6.240)}\\
&0.75 & \textbf{31.397(58.095)} & 51.404(64.778) & 63.834(152.047)    &   68.384(106.082)\\
&0.95 & 132.283\textbf{(166.269)} & \textbf{127.967}(250.006) & 214.595(559.565) &   282.834(646.127)\\
\hline

&0.05 &   \textbf{0.012(0.018)} & 0.190(0.105) & 0.207(0.116) & 0.051(0.075) \\
&0.25 &   \textbf{0.011(0.018)} & 0.048(0.111) & 0.045(0.092) & 0.057(0.079) \\
M3&0.50 & \textbf{0.012(0.018)} & 0.042(0.088) & 0.040(0.105) & 0.044(0.072) \\
&0.75 &   \textbf{0.014(0.019)} & 0.084(0.130) & 0.094(0.141) & 0.052(0.071) \\
&0.95 &   \textbf{0.017(0.023)} & 0.263(0.234) & 0.296(0.197) & 0.083(0.126) \\
\hline
\end{tabular}
\caption{ Mean squared errors (MSEs) and standard deviations (SDs) of the quantiles of 
samples generated by our conditional diffusion models, WGANs, Sinkhorn GANs, and k-NN. The smallest MSEs and SDs for each quantile are highlighted in bold.}
\label{quantile mse}
\end{table*}

\subsection{Test Statistic}

Conditional mutual information (CMI) $I(X;Y|Z)$ is defined as 
\begin{eqnarray*}
\iiint p_{X,Y,Z}(x,y,z)
\log \frac{p_{X,Y,Z}(x,y,z)}{p_{X,Z}(x,z)p_{Y|Z}(y|z)}dxdydz,
\end{eqnarray*}
where $p_{X,Y,Z}(x,y,z)$ is the joint density of $(X,Y,Z)$, $p_{X,Z}(x,z)$ is the joint density of $(X,Z)$, and $p_{Y|Z}(y|z)$ is the conditional density of $Y$ given $Z=z$. CMI as a tool for measuring conditional dependency, with $I(X;Y|Z)=0\iff X \perp \!\!\! \perp Y|Z$, has been  used in conditional independence testing \cite{runge2018conditional,li2024k}. Its major advantages include not needing the data to adhere to particular distributional assumptions or the features to have specific dependency relationships, making CMI applicable to a wide range of real-world datasets \cite{mukherjee2020ccmi}.


In Equation (\ref{CRT}), our test statistic $\bm{T}(\bm X,\bm Y,\bm Z)$ is set as 
  the  classifier-based CMI estimator (CCMI, \citeauthor{mukherjee2020ccmi}, \citeyear{mukherjee2020ccmi}). 
Specifically,  $I(X;Y|Z)$ can be  expressed  in terms of Kullback-Leibler (KL) divergence:
\begin{equation}\label{KLdiv}
I(X;Y|Z)=d_{\text{KL}}(p_{X,Y,Z}(x,y,z)||p_{X,Z}(x,z)p_{Y|Z}(y|z)),
\end{equation}
where $d_{\text{KL}}(f||g)$ denotes the KL divergence between two distribution functions $F$ and $G$, with density functions $f(x)$ and $g(x)$, respectively. We further utilize the Donsker-Varadhan (DV) representation of $d_{\text{KL}}(f||g)$, 
\begin{equation}\label{DV}
\sup_{s\in \mathcal{S}} \left[\mathbb{E}_{w\sim f}s(w)-\log{\mathbb{E}_{w\sim g}\exp\{s(w)\}}\right],
\end{equation}
where the function class $\mathcal{S}$ includes all functions with finite expectations. In fact, the optimal function in (\ref{DV}) is given by $s^{*}(x)=\log\{{f(x)}/{g(x)}\}$ \cite{belghazi2018mutual}, which leads to:
\begin{equation}\label{DVoptimal}
\begin{aligned}
d_{\text{KL}}(f||g)=\mathbb{E}_{w\sim f}\log &\left\{{f(w)}/{g(w)}\right\} \\
&-\log[\mathbb{E}_{w\sim g} \left\{{f(w)}/{g(w)}\right\}].
\end{aligned}
\end{equation}

Next, our primary goal is to empirically estimate   (\ref{DVoptimal}) with $f=p_{X,Y, Z}(x,y, z)$ and $g=p_{X,Z}(x,z)p_{Y|Z}(y|z)$, which requires samples from both $p_{X,Y, Z}(x,y, z)$ and $p_{X,Z}(x,z)p_{Y|Z}(y|z)$. 
Following the approach of \citeauthor{li2024k} \shortcite{li2024k}, we use the 1-NN sampling algorithm to estimate the conditional distribution of $Y|Z$. For further details, see Algorithm \ref{NNS} in Supplementary Materials.

Finally, we formalize the classifier-based CMI estimator; see Algorithm \ref{main:algorithm2} in Supplementary Materials. Specifically, consider a data set $V$ consisting of $2n$ i.i.d. samples $\{W_i:=(X_{i}, Y_{i}, Z_{i} )\}_{i=1}^{2n}$ with $(X_{i}, Y_{i}, Z_{i})\sim p_{X,Y,Z}(x,y,z)$. We divide $V$ into two parts, $V_1$ and $V_2$, each containing $n$ samples. Using Algorithm \ref{NNS}, we generate a new data set $V'$ with $n$ samples from $V_1$ and $V_2$. Assigning labels $l=1$ to all samples in $V_2$ (positive samples drawn from $p_{X,Y,Z}(x,y,z)$) and $l=0$ to all samples in $V'$ (negative samples drawn from $p_{X,Z}(x,z)p_{Y|Z}(y|z)$). In this supervised classification task, we train a binary classifier using advanced models like XGBoost \cite{sen2017model,chen2016xgboost} or deep neural networks \cite{goodfellow2016deep}. The classifier outputs the predicted probability $\alpha_{m}=P(l=1|W_m)$ for a given sample $W_m$, which leads to an estimator of the likelihood ratio on $W_m$ given by $\widehat{L}(W_m)= {\alpha_{m}}/(1-\alpha_{m})$. Based on Equations (\ref{KLdiv}) and (\ref{DVoptimal}), we can derive an estimator for $I(X;Y|Z)$, 
\begin{align}\label{MIE}
\widehat{I}(X;Y|Z)&:=\widehat{d}_{\text{KL}}(p_{X,Y,Z}(x,y,z)||p_{X,Z}(x,z)p_{Y|Z}(y|z))\nonumber\\
&={d}^{-1}    \sum _{i=1}^{d}\log \widehat{L}(W_{i}^{f})-\log \{{d}^{-1}\sum _{j=1}^{d}\widehat{L}(W_{j}^{g})\},
\end{align}
where $d=\lfloor n/3 \rfloor$ with  $\lfloor t\rfloor$ being the largest integer not greater than $t$, $W_{i}^{f}$ is a sample in  $V_f^{\mbox{test}}$,  and $W_{j}^{g}$ is a sample in $V_g^{\mbox{test}}$, with $V_f^{\mbox{test}}$ and $V_g^{\mbox{test}}$ defined in Algorithm \ref{main:algorithm2}.

\subsection{Computation of the $p$-value}
We calculate the $p$-value based on the  score-based conditional diffusion models to make informed decisions regarding the hypothesis testing. Specifically, 
by Algorithm \ref{algo:card sampler}, 
we  independently draw  pseudo samples $\widehat{X}_{i}^{(b)}\sim \widehat{p}(\cdot|Z_i)$ for each $i$ across $b=1, \ldots, B$, where $\widehat{p}(\cdot|Z_i)$ is the conditional density of $\widehat{X}$ given $Z_i$ and $B$ is the number of repetitions. Conditional on $\bm{Z}$,   all  $\widehat{\bm{X}}^{(b)}:=(\widehat{X}_{1}^{(b)}, \ldots, \widehat{X}_{n}^{(b)})^T$ are independent of
 $\bm{Y}$ and also $\bm{X}$. We denote the CMI estimator of $I({X};Y|Z)$  based on $(\widehat{\bm{X}}^{(b)}, \bm{Y}, \bm{Z})$ as  $\widehat{\mbox{CMI}}^{(b)}$  and denote the estimator based on $(\bm{X}, \bm{Y}, \bm{Z})$ as  $\widehat{\mbox{CMI}}$.
According to Theorem 1 in \citeauthor{mukherjee2020ccmi} \shortcite{mukherjee2020ccmi}, 
$\widehat{I}(X;Y|Z)$ is a consistent estimator of $I(X;Y|Z)$. We calculate the $p$-value using Equation (\ref{CRT}) by substituting $\bm{T}(\bm X^{(b)},\bm Y,\bm Z)$ and $\bm{T}(\bm X,\bm Y,\bm Z)$  with $\widehat{\mbox{CMI}}^{(b)}$ and $\widehat{\mbox{CMI}}$, respectively.
The pseudo code is summarized in Algorithm \ref{main:algorithm}. In the next section, we will prove  that our test asymptotically achieves a valid control of type I error. 

\begin{algorithm}[tb]
\caption{ Conditional diffusion models based conditional independence testing (CDCIT)}
\label{main:algorithm}
\textbf{Input}: Dataset $\mathcal{D}_{\text{T}}=({\bm{X}}, \bm{Y}, \bm{Z})$ consisting of $n$ i.i.d. samples from $p_{X,Y,Z}(x,y,z)$ and unlabelled dataset $\mathcal{D}_{\text{U}}=({\bm{X}}^{\text{U}}, \bm{Z}^{\text{U}})$ consisting of $N$ i.i.d. samples from $p_{X,Z}(x,z)$. \\
\textbf{Parameter}: The number of repetitions $B$; the significance level $\alpha$.\\
\textbf{Output}:  Accept $H_0:X \perp \!\!\! \perp Y|Z$ or $H_1:X \not \! \perp \!\!\! \perp Y|Z$.
\begin{algorithmic}[1] 
\STATE Use Algorithm \ref{main:algorithm2} to obtain $\widehat{\mbox{CMI}}$ based on $\mathcal{D}_{\text{T}}$.
\STATE Use Algorithm \ref{algo:train card} to obtain the score network $\widehat{s}$ based on $\mathcal{D}_{\text{U}}$.
\STATE $b=1$.
\WHILE{$b\leq B$}
\STATE For each $i\in \{ 1,\ldots,n \}$, given $Z_i$, produce $\widehat{X}_i^{(b)}$ using 
Algorithm \ref{algo:card sampler} with $Z_i$ and $\widehat{s}$ as input. Let $\widehat{\bm{X}}^{(b)}:=(\widehat{X}_{1}^{(b)}, \ldots, \widehat{X}_{n}^{(b)})^T$.
\STATE Use Algorithm \ref{main:algorithm2} to obtain $\widehat{\mbox{CMI}}{}^{(b)}$ based on $(\widehat{\bm{X}}^{(b)}, \bm{Y}, \bm{Z})$.
\STATE $b=b+1$.
\ENDWHILE
\STATE Compute $p$-value: $p:=\big[1+\sum_{b = 1}^{B}\bm{\mbox{I}}\big\{\widehat{\mbox{CMI}}{}^{(b)}\geq \widehat{\mbox{CMI}}\big\} \big]/(1+B)$.
\IF {$p\geq \alpha$}
\STATE Accept $H_0:X \perp \!\!\! \perp Y|Z$.
\ELSE
\STATE Accept $H_1:X \not \! \perp \!\!\! \perp Y|Z$.
\ENDIF
\end{algorithmic}
\end{algorithm}

\section{Theoretical Results}
In this section, we present our main theoretical results, with  
all  proofs 
deferred to Supplementary Materials. 
Denote ${p}^{(n)}(\cdot|\bm Z):=
\prod_{i=1}^n
{p}(\cdot|Z_i)$ and $\widehat{p}^{(n)}(\cdot|\bm Z):=\prod_{i=1}^n\widehat{p}(\cdot|Z_i)$.
In Theorem \ref{thm2}, we bound the excess type I error conditionally on $\bm{Y}$ and $\bm{Z}$  by the total variation distance between $\widehat{p}^{(n)}(\cdot|\bm{Z})$ and $p^{(n)}(\cdot|\bm{Z})$.
\begin{thm}\label{thm2}
Assume $H_0:X \perp \!\!\! \perp Y|Z$ is true. For any significance level $\alpha \in (0,1)$, the $p$-value obtained from Algorithm \ref{main:algorithm} satisfies
\begin{equation*}
    P(p\leq \alpha |\bm{Y},\bm{Z}) \leq \alpha+d_{\text{TV}}(p^{(n)}(\cdot|\bm{Z}),\widehat{p}^{(n)}(\cdot|\bm{Z}) ).
\end{equation*}
\end{thm}
An immediate implication of Theorem \ref{thm2} is that  the type I error rate  can be unconditionally controlled as
\begin{equation*}
    P(p\leq \alpha |H_0) \leq \alpha+\mathbb{E}[d_{\text{TV}}(p^{(n)}(\cdot|\bm Z),\widehat{p}^{(n)}(\cdot|\bm Z))].
\end{equation*}
Then applying Theorem \ref{thm1}
to this inequality yields
the following
Corollary~\ref{corollary}, which shows that our CI testing procedure 
can asymptotically control the
 type I error
 at level~$\alpha$. 
\begin{cor} \label{corollary}
Assume $n\cdot N^ {-\Gamma_1(k,\alpha)} \cdot \left(\log N \right) ^{\Gamma_2(k,\alpha)}=o(1)$. Under assumptions in Theorems \ref{thm1} and \ref{thm2}, the $p$-value obtained from Algorithm \ref{main:algorithm} satisfies
\begin{equation*}
    P(p\leq \alpha |H_0) \leq \alpha+o(1).
\end{equation*}
\end{cor} 
\begin{remark}
Since $\log N=O(N^{\delta})$ for any $\delta>0$, the sample size assumption in Corollary \ref{corollary} implies that the  sample size $N$ in $\mathcal{D}_{\text{U}}$ needs to satisfy $N\gg n^{1/(\Gamma_1(k,\alpha)-\delta)}$ for any sufficiently small $\delta>0$ in order to asymptotically control the type I error. 
\end{remark} 


\section{Synthetic Data Analysis}
We evaluate our method, CDCIT, on  synthetic datasets, and  compare it with the seven state-of-the-art
(SOTA) methods: GCIT \cite{bellot2019conditional}, NNSCIT \cite{li2023nearest}, the classifier-based CI test (CCIT)  \cite{sen2017model}, the kernel-based CI test (KCIT) \cite{zhang2011kernel}, LPCIT \cite{scetbon2022asymptotic}, DGCIT \cite{shi2021double}, and NNLSCIT \cite{li2024k}. We 
set the number of repetitions $B $ to $ 100$ and the significance level $\alpha$ to $0.05$. We  report the type I error rate and the testing power under $H_1$ for all methods in each experiment. All the results are presented as an average over $100$ independent trials. 
We  present additional simulation studies, the detailed training parameter settings   for our CDCIT and the real data
analysis in Supplementary Materials. The  source code of our CDCIT is publicly available at:

\begin{links}
\link{Code}{https://github.com/Yanfeng-Yang-0316/CDCIT}
\end{links}


\begin{figure*}[ht!]
    \centering
    \begin{minipage}{1\linewidth}
        \centering
        \includegraphics[width=1\linewidth]{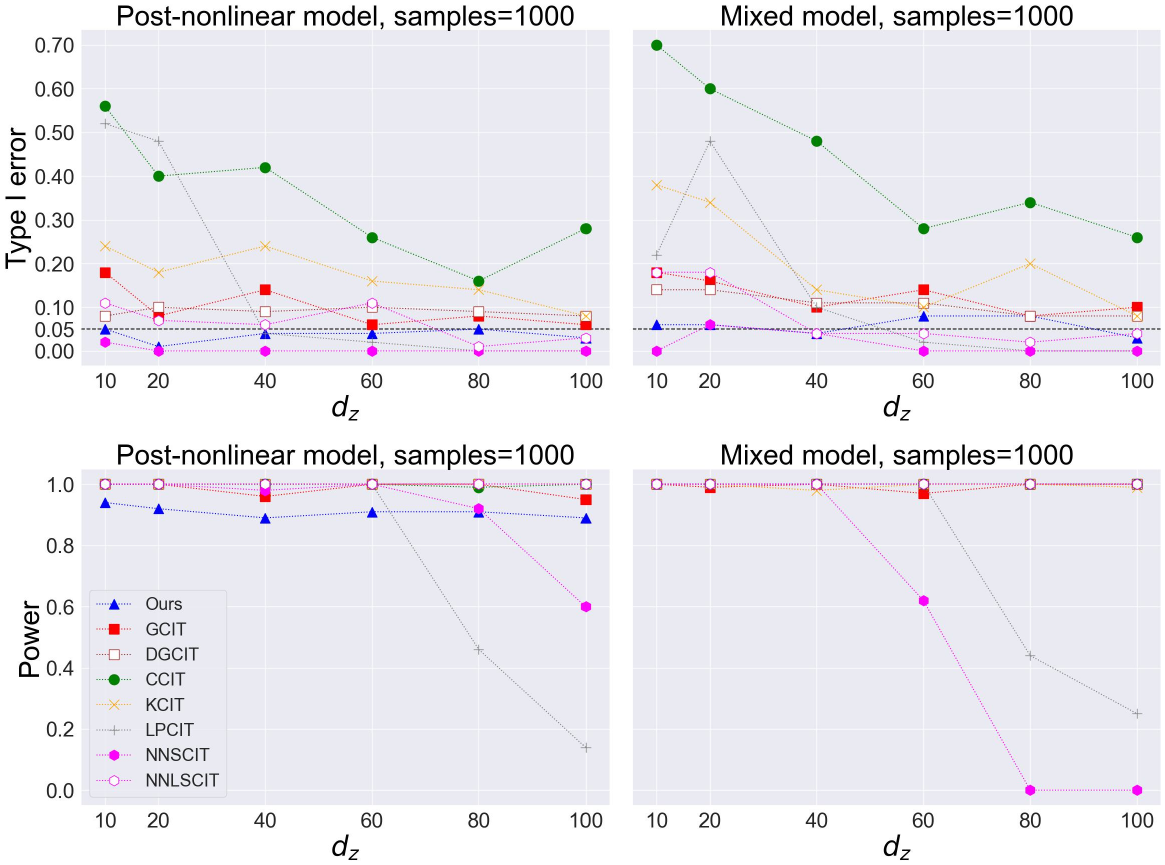}
    \end{minipage}
    
    \caption{Comparison of the type I error (lower is better) and power (higher is better) of our method
with seven SOTA methods on the post-nonlinear model (\ref{fork_structue}) and mixed model (\ref{dis_and_continuous})
with varying dimension of $Z$. Under the mixed model, the power of our method, as well as those of DGCIT, CCIT, and NNLSCIT, stays consistently at $1$ across different $d_z$.}
    \label{simulation_fork}
 \end{figure*}
\textbf{Scenario I: the post-nonlinear model.} The first synthetic dataset is generated using the post-nonlinear model similar to those in \citeauthor{zhang2011kernel} \shortcite{zhang2011kernel}, \citeauthor{bellot2019conditional} \shortcite{bellot2019conditional}, \citeauthor{scetbon2022asymptotic} \shortcite{scetbon2022asymptotic},  and \citet{li2024k}. Specifically, the triples $(X, Y, Z)$ under $H_0$ and $H_1$ are generated using the following models: 
\begin{align}\label{fork_structue}
    H_0: & X=f_{1}(\overline{Z}+0.25\cdot\epsilon_{x}),\\ 
    & Y=f_{2}(\overline{Z}+0.25\cdot\epsilon_{y}), \nonumber \\
     H_1: & X=f_{1}(\overline{Z}+0.25\cdot\epsilon_{x})+0.5\cdot\epsilon_{b} , \nonumber \\
     & Y= f_{2}(\overline{Z}+0.25\cdot\epsilon_{y})+0.5\cdot\epsilon_{b}, \nonumber
\end{align}
where $\overline{Z}$ is the sample mean of $Z=(z_{1},\ldots,z_{d_z})$, all $z_l$ in $Z$, $\epsilon_{x}$, $\epsilon_{y}$ and $\epsilon_{b}$ are i.i.d. samples generated from the standard Gaussian distribution, 
and functions $f_1$ and $f_2$ are randomly sampled from the set $\{ x, x^{2}, x^{3}, \tanh(x), \cos(x)\}$, and $d_z$ represents the dimension of $Z$. 

\textbf{Scenario II: the mixed continuous and discrete conditioning-set  model.} 
The conditioning variable set \( Z = (z_1, \ldots, z_{d_z}) \) is mixed-type, consisting of \(\lfloor d_z/2 \rfloor \) continuous variables \((z_1, \ldots, z_{\lfloor d_z/2 \rfloor})\) and \(d_z- \lfloor d_z/2 \rfloor \) discrete variables \((z_{\lfloor d_z/2 \rfloor + 1}, \ldots, z_{d_z})\). We use only \((z_1, z_2, \ldots, \lfloor 2 \cdot d_z / 3 \rfloor)\) to generate \( X \) and \( Y \) under both \( H_0 \) and \( H_1 \) in the true model.
 Specifically, 
\begin{align}\label{dis_and_continuous}
    H_0: & X=\dfrac{1}{\lfloor 2 \cdot d_z /3 \rfloor} \sum_{i=1}^{\lfloor 2 \cdot d_z /3 \rfloor} z_i +0.33 \cdot \epsilon_x,\\ 
    & Y=\dfrac{1}{\lfloor 2 \cdot d_z /3 \rfloor} \sum_{i=1}^{\lfloor 2 \cdot d_z /3 \rfloor} z_i +0.33 \cdot \epsilon_y, \nonumber \\
     H_1: & X=\dfrac{1}{\lfloor 2 \cdot d_z /3 \rfloor} \sum_{i=1}^{\lfloor 2 \cdot d_z /3 \rfloor} z_i +0.33 \cdot \epsilon_{b} , \nonumber \\
     & Y= \dfrac{1}{\lfloor 2 \cdot d_z /3 \rfloor} \sum_{i=1}^{\lfloor 2 \cdot d_z /3 \rfloor} z_i +0.33 \cdot \epsilon_{b}, \nonumber
\end{align}
where $z_1,\ldots,z_{\lfloor d_z/2 \rfloor} \mathop{\sim}\limits_{}^{i.i.d.} N(0,1),~ z_{\lfloor d_z/2 \rfloor+1}, \ldots, z_{d_z} \mathop{\sim}\limits_{}^{i.i.d.} \\ 2 \cdot \text{Bernoulli}(1,0.5)-1$, and $\epsilon_x$, $\epsilon_y$ and $\epsilon_b$ all follow the standard Gaussian distribution.

For each experiment, 1000 samples are generated. We use $N=500$ to train the conditional sampler and $n=500$ to compute the test statistic in our CDCIT. We vary $d_z$, the dimension of $Z$, from 10 to 100. The results are shown in Figure \ref{simulation_fork}. More results regarding $N$ and $n$ are provided in Figures \ref{simulation_fork_dz20} and \ref{different_N_n_B} in Supplementary Materials.

We have the following observations. First, in both post-nonlinear and mixed models, our test controls type I error very well  and achieves high power under $H_1$ as $d_z$  increases. Second, NNSCIT has satisfactory performance in controlling type I error, but it loses power under $H_1$, especially when $d_z$  exceeds 40 in the mixed model. Third, although CCIT, KCIT and GCIT have adequate power under $H_1$, they have inflated type I errors in almost all scenarios. Fourth, DGCIT and NNLSCIT sometimes fail to control the type I error well, especially when  $d_z\leq 20$. 
Fifth, LPCIT shows weak performance on both type I error and testing power.

Figure \ref{computing_time_second} in Supplementary Materials reports the timing performance of all considered methods for a single test. Our CDCIT is found to be highly computationally efficient even when dealing with large sample sizes and high-dimensional conditioning sets.

\section{Conclusion}
We introduce a novel CI testing procedure using the conditional diffusion models to approximate the distribution of $X|Z$. 
We have theoretically and empirically shown that the distribution of the generated samples is very close to the true conditional distribution. We use a computationally efficient classifier-based CMI estimator as the test statistic, which captures intricate dependence structures among variables. We demonstrate that our proposed test achieves valid control of the type I and type II errors. Furthermore, our test remains highly computationally efficient, even when dealing with high-dimensional conditioning sets. 
Our method has the potential to broaden the applicability of causal discovery in real-world scenarios, such as gene regulatory network, the identification of disease-associated genes,  and intricate social networks, thereby aiding in the identification of relationships and patterns within complex systems. 
\clearpage
\section*{Acknowledgments}
Dr. Ziqi Chen’s work was partially supported by National
Natural Science Foundation of China (NSFC) (12271167
and 72331005) and Basic Research Project of Shanghai Science and Technology Commission (22JC1400800). We thank the anonymous reviewers for their helpful
comments.


\bibliography{aaai25}

\begin{thebibliography}{64}
\providecommand{\natexlab}[1]{#1}

\bibitem[{Arjovsky, Chintala, and Bottou(2017)}]{arjovsky2017wasserstein}
Arjovsky, M.; Chintala, S.; and Bottou, L. 2017.
\newblock Wasserstein generative adversarial networks.
\newblock In \emph{International Conference on Machine Learning}, 214--223.

\bibitem[{Beca et~al.(2015)Beca, Pereira, Cameselle-Teijeiro et~al.}]{PPP2R2A_breast_cancer}
Beca, F.; Pereira, M.~T.; Cameselle-Teijeiro, J.~F.; et~al. 2015.
\newblock Altered PPP2R2A and Cyclin D1 expression defines a subgroup of aggressive luminal-like breast cancer.
\newblock \emph{BMC Cancer}, 15(1): 285.

\bibitem[{Belghazi et~al.(2018)Belghazi, Baratin, Rajeshwar, Ozair, Bengio, Courville et~al.}]{belghazi2018mutual}
Belghazi, M.~I.; Baratin, A.; Rajeshwar, S.; Ozair, S.; Bengio, Y.; Courville, A.; et~al. 2018.
\newblock Mutual information neural estimation.
\newblock In \emph{International Conference on Machine Learning}, 531--540.

\bibitem[{Bellot and van~der Schaar(2019)}]{bellot2019conditional}
Bellot, A.; and van~der Schaar, M. 2019.
\newblock Conditional independence testing using generative adversarial networks.
\newblock In \emph{Advances in Neural Information Processing Systems}, volume~32.

\bibitem[{Berrett et~al.(2020)Berrett, Wang, Barber, and Samworth}]{berrett2020conditional}
Berrett, T.~B.; Wang, Y.; Barber, R.~F.; and Samworth, R.~J. 2020.
\newblock The conditional permutation test for independence while controlling for confounders.
\newblock \emph{Journal of the Royal Statistical Society: Series B (Statistical Methodology)}, 82(1): 175--197.

\bibitem[{B{\o}rresen-Dale(2003)}]{TP53_breast_cancer}
B{\o}rresen-Dale, A.-L. 2003.
\newblock TP53 and breast cancer.
\newblock \emph{Human Mutation}, 21(3): 292--300.

\bibitem[{Cand{\`e}s et~al.(2018)Cand{\`e}s, Fan, Janson, and Lv}]{candes2018panning}
Cand{\`e}s, E.; Fan, Y.; Janson, L.; and Lv, J. 2018.
\newblock Panning for gold:`model-X' knockoffs for high dimensional controlled variable selection.
\newblock \emph{Journal of the Royal Statistical Society: Series B (Statistical Methodology)}, 80(3): 551--577.

\bibitem[{Chen and Guestrin(2016)}]{chen2016xgboost}
Chen, T.; and Guestrin, C. 2016.
\newblock Xgboost: A scalable tree boosting system.
\newblock In \emph{ACM SIGKDD International Conference on Knowledge Discovery and Data Mining}, 785--794.

\bibitem[{Cover and Thomas(2012)}]{cover2012elements}
Cover, T.~M.; and Thomas, J.~A. 2012.
\newblock \emph{Elements of Information Theory}.
\newblock John Wiley \& Sons.

\bibitem[{Curtis et~al.(2012)Curtis, Shah, Chin, Turashvili, Rueda, Dunning et~al.}]{nature_breast_cancer}
Curtis, C.; Shah, S.; Chin, S.-F.; Turashvili, G.; Rueda, O.; Dunning, M.; et~al. 2012.
\newblock The genomic and transcriptomic architecture of 2,000 breast tumors reveals novel subgroups.
\newblock \emph{Nature}, 486: 346–352.

\bibitem[{Dai, Shen, and Pan(2022)}]{dai2022significance}
Dai, B.; Shen, X.; and Pan, W. 2022.
\newblock Significance tests of feature relevance for a black-box learner.
\newblock \emph{IEEE Transactions on Neural Networks and Learning Systems}.

\bibitem[{Dai et~al.(2024)Dai, Ng, Luo, Spirtes, Stojanov, and Zhang}]{dai2024gene}
Dai, H.; Ng, I.; Luo, G.; Spirtes, P.; Stojanov, P.; and Zhang, K. 2024.
\newblock Gene Regulatory Network Inference in the Presence of Dropouts: a Causal View.
\newblock arXiv:2403.15500.

\bibitem[{Dhariwal and Nichol(2021)}]{dhariwal2021diffusion}
Dhariwal, P.; and Nichol, A. 2021.
\newblock Diffusion models beat gans on image synthesis.
\newblock In \emph{Advances in Neural Information Processing Systems}, volume~34.

\bibitem[{Doran et~al.(2014)Doran, Muandet, Zhang, and Sch{\"o}lkopf}]{doran2014permutation}
Doran, G.; Muandet, K.; Zhang, K.; and Sch{\"o}lkopf, B. 2014.
\newblock A permutation-based kernel conditional independence test.
\newblock In \emph{Conference on Uncertainty in Artificial Intelligence}, 132--141.

\bibitem[{Efron(2004)}]{efron2004selection}
Efron, B. 2004.
\newblock {Selection and estimation for large-scale simultaneous inference}.
\newblock Technical Report 2005-18B/232, Division of Biostatistics, Stanford University.

\bibitem[{Fu et~al.(2024)Fu, Yang, Wang, and Chen}]{Fu2024UnveilCD}
Fu, H.; Yang, Z.; Wang, M.; and Chen, M. 2024.
\newblock Unveil conditional diffusion models with classifier-free guidance: A sharp statistical theory.
\newblock arXiv:2403.11968.

\bibitem[{Fukumizu et~al.(2007)Fukumizu, Gretton, Sun, and Sch{\"o}lkopf}]{fukumizu2007kernel}
Fukumizu, K.; Gretton, A.; Sun, X.; and Sch{\"o}lkopf, B. 2007.
\newblock Kernel measures of conditional dependence.
\newblock In \emph{Advances in Neural Information Processing Systems}, volume~20.

\bibitem[{Genevay, Peyr{\'e}, and Cuturi(2018)}]{genevay2018learning}
Genevay, A.; Peyr{\'e}, G.; and Cuturi, M. 2018.
\newblock Learning generative models with sinkhorn divergences.
\newblock In \emph{International Conference on Artificial Intelligence and Statistics}, 1608--1617.

\bibitem[{Goodfellow, Bengio, and Courville(2016)}]{goodfellow2016deep}
Goodfellow, I.; Bengio, Y.; and Courville, A. 2016.
\newblock \emph{Deep learning}.
\newblock MIT Press.

\bibitem[{Hall, Racine, and Li(2004)}]{hall2004cross}
Hall, P.; Racine, J.; and Li, Q. 2004.
\newblock Cross-validation and the estimation of conditional probability densities.
\newblock \emph{Journal of the American Statistical Association}, 99(468): 1015--1026.

\bibitem[{Hall and Yao(2005)}]{hall2005approximating}
Hall, P.; and Yao, Q. 2005.
\newblock Approximating conditional distribution functions using dimension reduction.
\newblock \emph{The Annals of Statistics}, 1404--1421.

\bibitem[{Han, Zheng, and Zhou(2022)}]{han2022card}
Han, X.; Zheng, H.; and Zhou, M. 2022.
\newblock Card: Classification and regression diffusion models.
\newblock In \emph{Advances in Neural Information Processing Systems}, volume~35.

\bibitem[{He, Zhang, and Long(2016)}]{APC_breast_cancer}
He, K.; Zhang, L.; and Long, X. 2016.
\newblock Quantitative assessment of the association between APC promoter methylation and breast cancer.
\newblock \emph{Oncotarget}, 7(25): 37920--37930.

\bibitem[{Ho, Jain, and Abbeel(2020)}]{ho2020denoising}
Ho, J.; Jain, A.; and Abbeel, P. 2020.
\newblock Denoising diffusion probabilistic models.
\newblock In \emph{Advances in Neural Information Processing Systems}, volume~33.

\bibitem[{Ho et~al.(2022)Ho, Salimans, Gritsenko et~al.}]{ho2022video}
Ho, J.; Salimans, T.; Gritsenko, A.; et~al. 2022.
\newblock Video diffusion models.
\newblock In \emph{Advances in Neural Information Processing Systems}, volume~35.

\bibitem[{Izbicki and Lee(2017)}]{izbicki2017converting}
Izbicki, R.; and Lee, A.~B. 2017.
\newblock Converting high-dimensional regression to high-dimensional conditional density estimation.
\newblock \emph{Electronic Journal of Statistics}, 11: 2800--2831.

\bibitem[{Khera and Kathiresan(2017)}]{khera2017genetics}
Khera, A.~V.; and Kathiresan, S. 2017.
\newblock Genetics of coronary artery disease: discovery, biology and clinical translation.
\newblock \emph{Nature Reviews Genetics}, 18(6): 331--344.

\bibitem[{Koller and Friedman(2009)}]{koller2009probabilistic}
Koller, D.; and Friedman, N. 2009.
\newblock \emph{Probabilistic graphical models: principles and techniques}.
\newblock MIT Press.

\bibitem[{Kuchibhotla(2020)}]{kuchibhotla2020exchangeability}
Kuchibhotla, A.~K. 2020.
\newblock Exchangeability, conformal prediction, and rank tests.
\newblock arXiv:2005.06095.

\bibitem[{Lauritzen(1996)}]{lauritzen1996graphical}
Lauritzen, S.~L. 1996.
\newblock \emph{Graphical models}, volume~17.
\newblock Clarendon Press.

\bibitem[{Li and Fan(2020)}]{li2020nonparametric}
Li, C.; and Fan, X. 2020.
\newblock On nonparametric conditional independence tests for continuous variables.
\newblock \emph{Wiley Interdisciplinary Reviews: Computational Statistics}, 12(3): e1489.

\bibitem[{Li et~al.(2023)Li, Chen, Zhu, Wang, and Wen}]{li2023nearest}
Li, S.; Chen, Z.; Zhu, H.; Wang, C.; and Wen, W. 2023.
\newblock Nearest-neighbor sampling based conditional independence testing.
\newblock In \emph{AAAI Conference on Artificial Intelligence}, volume~37, 8631--8639.

\bibitem[{Li et~al.(2024)Li, Zhang, Zhu, Wang, Shu, Chen et~al.}]{li2024k}
Li, S.; Zhang, Y.; Zhu, H.; Wang, C.; Shu, H.; Chen, Z.; et~al. 2024.
\newblock K-nearest-neighbor local sampling based conditional independence testing.
\newblock In \emph{Advances in Neural Information Processing Systems}, volume~36.

\bibitem[{Liu et~al.(2022)Liu, Katsevich, Janson, and Ramdas}]{liu2022fast}
Liu, M.; Katsevich, E.; Janson, L.; and Ramdas, A. 2022.
\newblock Fast and powerful conditional randomization testing via distillation.
\newblock \emph{Biometrika}, 109(2): 277--293.

\bibitem[{Maguire et~al.(2015)Maguire, Leonidou, Wai, Marchiò, Ng, Sapino et~al.}]{SF3B1_breast_cancer}
Maguire, S.; Leonidou, A.; Wai, P.; Marchiò, C.; Ng, C.; Sapino, A.; et~al. 2015.
\newblock SF3B1 mutations constitute a novel therapeutic target in breast cancer.
\newblock \emph{The Journal of Pathology}, 235(4): 571--580.

\bibitem[{Mei et~al.(2024)Mei, Mei, Chang, Liu, Zhou, Zhu et~al.}]{FOXO3_breast_cancer}
Mei, W.; Mei, B.; Chang, J.; Liu, Y.; Zhou, Y.; Zhu, N.; et~al. 2024.
\newblock Role and regulation of FOXO3a: new insights into breast cancer therapy.
\newblock \emph{Frontiers in Pharmacology}, 15.

\bibitem[{Mesner and Shalizi(2020)}]{mesner2020conditional}
Mesner, O.~C.; and Shalizi, C.~R. 2020.
\newblock Conditional mutual information estimation for mixed, discrete and continuous data.
\newblock \emph{IEEE Transactions on Information Theory}, 67(1): 464--484.

\bibitem[{Mooij and Heskes(2013)}]{mooij2013cyclic}
Mooij, J.~M.; and Heskes, T. 2013.
\newblock Cyclic causal discovery from continuous equilibrium data.
\newblock In \emph{Conference on Uncertainty in Artificial Intelligence}, 431--439.

\bibitem[{Mukherjee, Asnani, and Kannan(2020)}]{mukherjee2020ccmi}
Mukherjee, S.; Asnani, H.; and Kannan, S. 2020.
\newblock CCMI: Classifier based conditional mutual information estimation.
\newblock In \emph{Conference on Uncertainty in Artificial Intelligence}, 1083--1093.

\bibitem[{Ng, Ghassami, and Zhang(2020)}]{ng2020role}
Ng, I.; Ghassami, A.; and Zhang, K. 2020.
\newblock On the role of sparsity and dag constraints for learning linear dags.
\newblock In \emph{Advances in Neural Information Processing Systems}, volume~33.

\bibitem[{Pearl(1988)}]{pearl1988probabilistic}
Pearl, J. 1988.
\newblock \emph{Probabilistic reasoning in intelligent systems: networks of plausible inference}.
\newblock Morgan kaufmann.

\bibitem[{Pereira et~al.(2016)Pereira, Chin, Rueda et~al.}]{MLLT4_breast_cancer}
Pereira, B.; Chin, S.-F.; Rueda, O.; et~al. 2016.
\newblock Erratum: The somatic mutation profiles of 2,433 breast cancers refine their genomic and transcriptomic landscapes.
\newblock \emph{Nature Communications}, 7(11479): 11908.

\bibitem[{Runge(2018)}]{runge2018conditional}
Runge, J. 2018.
\newblock Conditional independence testing based on a nearest-neighbor estimator of conditional mutual information.
\newblock In \emph{International Conference on Artificial Intelligence and Statistics}, 938--947.

\bibitem[{Sachs et~al.(2005)Sachs, Perez, Pe'er, Lauffenburger, and Nolan}]{sachs2005causal}
Sachs, K.; Perez, O.; Pe'er, D.; Lauffenburger, D.~A.; and Nolan, G.~P. 2005.
\newblock Causal protein-signaling networks derived from multiparameter single-cell data.
\newblock \emph{Science}, 308(5721): 523--529.

\bibitem[{Scetbon, Meunier, and Romano(2022)}]{scetbon2022asymptotic}
Scetbon, M.; Meunier, L.; and Romano, Y. 2022.
\newblock An asymptotic test for conditional independence using analytic kernel embeddings.
\newblock In \emph{International Conference on Machine Learning}, 19328--19346.

\bibitem[{Sen et~al.(2017)Sen, Suresh, Shanmugam, Dimakis, and Shakkottai}]{sen2017model}
Sen, R.; Suresh, A.~T.; Shanmugam, K.; Dimakis, A.~G.; and Shakkottai, S. 2017.
\newblock Model-powered conditional independence test.
\newblock In \emph{Advances in Neural Information Processing Systems}, volume~30.

\bibitem[{Shi et~al.(2021)Shi, Xu, Bergsma, and Li}]{shi2021double}
Shi, C.; Xu, T.; Bergsma, W.; and Li, L. 2021.
\newblock Double generative adversarial networks for conditional independence testing.
\newblock \emph{Journal of Machine Learning Research}, 22(285): 1--32.

\bibitem[{Shigekawa et~al.(2011)Shigekawa, Ijichi, Ikeda et~al.}]{FOXP1_breast_cancer}
Shigekawa, T.; Ijichi, N.; Ikeda, K.; et~al. 2011.
\newblock FOXP1, an Estrogen-Inducible Transcription Factor, Modulates Cell Proliferation in Breast Cancer Cells and 5-Year Recurrence-Free Survival of Patients with Tamoxifen-Treated Breast Cancer.
\newblock \emph{Hormones \& Cancer}, 2(5): 286--297.

\bibitem[{Song, Meng, and Ermon(2021)}]{songdenoising}
Song, J.; Meng, C.; and Ermon, S. 2021.
\newblock Denoising diffusion implicit models.
\newblock In \emph{International Conference on Learning Representations}.

\bibitem[{Song et~al.(2021)Song, Sohl-Dickstein, Kingma, Kumar, Ermon, and Poole}]{song2021score}
Song, Y.; Sohl-Dickstein, J.; Kingma, D.~P.; Kumar, A.; Ermon, S.; and Poole, B. 2021.
\newblock Score-based generative modeling through stochastic differential equations.
\newblock In \emph{International Conference on Learning Representations}.

\bibitem[{Spirtes, Glymour, and Scheines(2000)}]{spirtes2000causation}
Spirtes, P.; Glymour, C.~N.; and Scheines, R. 2000.
\newblock \emph{Causation, prediction, and search}.
\newblock MIT Press.

\bibitem[{Su and White(2008)}]{su2008nonparametric}
Su, L.; and White, H. 2008.
\newblock A nonparametric Hellinger metric test for conditional independence.
\newblock \emph{Econometric Theory}, 24(4): 829--864.

\bibitem[{Su and White(2014)}]{su2014testing}
Su, L.; and White, H. 2014.
\newblock Testing conditional independence via empirical likelihood.
\newblock \emph{Journal of Econometrics}, 182(1): 27--44.

\bibitem[{Sun and Cai(2007)}]{sun2007oracle}
Sun, W.; and Cai, T.~T. 2007.
\newblock Oracle and adaptive compound decision rules for false discovery rate control.
\newblock \emph{Journal of the American Statistical Association}, 102(479): 901--912.

\bibitem[{Tsybakov(2009)}]{tsybakov2009nonparametric}
Tsybakov, A.~B. 2009.
\newblock \emph{Introduction to Nonparametric Estimation}.
\newblock New York: Springer.

\bibitem[{Verma et~al.(2020)Verma, Bakshi, Sharma, Sharma, Shah, Bhat et~al.}]{DNAH11_breast_cancer}
Verma, S.; Bakshi, D.; Sharma, V.; Sharma, I.; Shah, R.; Bhat, A.; et~al. 2020.
\newblock Genetic variants of 11 and 2 genes and their association with ovarian and breast cancer.
\newblock \emph{International Journal of Gynecology \& Obstetrics}, 148(1): 118--122.

\bibitem[{Wang et~al.(2015)Wang, Pan, Hu, Tian, and Zhang}]{wang2015conditional}
Wang, X.; Pan, W.; Hu, W.; Tian, Y.; and Zhang, H. 2015.
\newblock Conditional distance correlation.
\newblock \emph{Journal of the American Statistical Association}, 110(512): 1726--1734.

\bibitem[{Weglarczyk(2018)}]{kde_plot}
Weglarczyk, S. 2018.
\newblock Kernel density estimation and its application.
\newblock \emph{ITM Web of Conferences}, 23: 00037.

\bibitem[{Xiong et~al.(2022)Xiong, Chen, Lin et~al.}]{NR3C1_breast_cancer}
Xiong, H.; Chen, Z.; Lin, B.; et~al. 2022.
\newblock Naringenin Regulates FKBP4/NR3C1/NRF2 Axis in Autophagy and Proliferation of Breast Cancer and Differentiation and Maturation of Dendritic Cell.
\newblock \emph{Frontiers in Immunology}, 12.

\bibitem[{Yang et~al.(2023)Yang, Zhang, Song et~al.}]{yang2023diffusion}
Yang, L.; Zhang, Z.; Song, Y.; et~al. 2023.
\newblock Diffusion models: A comprehensive survey of methods and applications.
\newblock \emph{ACM Computing Surveys}, 56(4): 1--39.

\bibitem[{Zan et~al.(2022)Zan, Meynaoui, Assaad et~al.}]{zan2022conditional}
Zan, L.; Meynaoui, A.; Assaad, C.~K.; et~al. 2022.
\newblock A conditional mutual information estimator for mixed data and an associated conditional independence test.
\newblock \emph{Entropy}, 24(9): 1234.

\bibitem[{Zhang et~al.(2011)Zhang, Peters, Janzing, and Sch{\"o}lkopf}]{zhang2011kernel}
Zhang, K.; Peters, J.; Janzing, D.; and Sch{\"o}lkopf, B. 2011.
\newblock Kernel-based conditional independence test and application in causal discovery.
\newblock In \emph{Conference on Uncertainty in Artificial Intelligence}, 804--813.

\bibitem[{Zhu, Ng, and Chen(2020)}]{zhucausal}
Zhu, S.; Ng, I.; and Chen, Z. 2020.
\newblock Causal Discovery with Reinforcement Learning.
\newblock In \emph{International Conference on Learning Representations}.

\bibitem[{Zhu et~al.(2018)Zhu, Zheng, Zhang et~al.}]{zhu2018causal}
Zhu, Z.; Zheng, Z.; Zhang, F.; et~al. 2018.
\newblock Causal associations between risk factors and common diseases inferred from GWAS summary data.
\newblock \emph{Nature Communications}, 9(1): 1--12.

\end{thebibliography}
 \clearpage

\appendix

\section*{\centering Supplementary Materials for ``Conditional Diffusion Models Based Conditional Independence Testing"}

\section{S1. Assumptions in Theoretical Guarantee for Sampling Quality}
We provide some necessary assumptions  for deriving Theorem \ref{thm1}.

\noindent{\bf Definition 1 (Hölder norm).} 
Let $k \geq 0$ be an integer and $\alpha \in (0,1]$. Given a function $v(x): \mathbb{R}^{d} \rightarrow \mathbb{R}$. Let $\mathbf{s}=(s_1,\ldots, s_d)$, $s_i \geq 0$ be integers, and $|\mathbf{s}|=\sum_{i=1}^d s_i$. Partial derivative $\partial^{\mathbf{s}} v(x)=\partial^{|\mathbf{s}|}v(x) / \partial x_1^{s_1} \ldots \partial x_d^{s_d}$.
The Hölder norm of $v(x)$ is defined as:
\begin{equation*}
\begin{array}{l}
    \|v(x) \|_{\mathcal{H}^{k,\alpha}(\mathbb{R}^{d})} \\  \\ =\max_{|\mathbf{s}|< k} \sup_{x \in \mathbb{R}^{d}}  |\partial^{\mathbf{s}} v(x)| \\ \\ 
    ~~~~+ \max_{|\mathbf{s}| =k} \sup_{x_1 \neq x_2} \dfrac{|\partial^{\mathbf{s}} v(x_1) -\partial^{\mathbf{s}} v(x_2)|}{\|x_1-x_2\|_{\infty}^{\alpha}}.
\end{array}
\end{equation*}

\noindent{\textbf{Assumption 1.} (Assumptions on $p_{X|Z}$).} 
Let $B_1,B_2,B_3$ be three positive constants. Assume that $\forall (x,z) \in \mathbb{R}^{d_x} \times \mathbb{R}^{d_z}$, $p(x|z)=\exp(-B_1\| x \|_2^2/2) \cdot v(x,z)$, where $v(x,z) \geq B_2$ and $\|v(x,z) \|_{\mathcal{H}^{k,\alpha}(\mathbb{R}^{d_x} \times \mathbb{R}^{d_z})} \leq B_3$. 

\noindent{\textbf{Assumption 2.} (Assumption on $p_{Z}$).} 
Let $B_4$ be a positive constant. Assume that $ \forall z \in \mathbb{R}^{d_z}, p_{Z}(z) \leq \exp(- B_4 \|z \|_2^2/2)$, where $p_{Z}(\cdot)$ is the density of $Z$.



Assumption 1 specifies the form of conditional density function. The bounded Hölder norm $\|v(x,z) \|_{\mathcal{H}^{k,\alpha}(\mathbb{R}^{d_x} \times \mathbb{R}^{d_z})} \leq B_3$ implies the smoothness of the conditional density and an upper bound of $v(x,z)$. 
Here, $k+\alpha$ determines the smoothness of $v(x,z)$. A larger $k+\alpha$ indicates a smoother $v(x,z)$, which facilitates the approximation of the conditional density function. This is consistent with Theorem \ref{thm1}, as a larger $k+\alpha$ yields a larger $\Gamma_1(k, \alpha)$, leading to faster convergence rate. The upper bound assumption of $v(x,z)$ is often required for effective density estimation \cite{tsybakov2009nonparametric}, and the lower bound of $v(x,z)$ plays a role in preventing the exploding of score function $\nabla \log p_{t}(\cdot|z)$. Both of them are required in \citeauthor{Fu2024UnveilCD} (\citeyear{Fu2024UnveilCD}) to analyse the approximation error between $\widehat{s}(x,z,t)$ and the real score $\nabla \log p_{t}(\cdot|z)$. This is the key to Lemma D.7 in \citeauthor{Fu2024UnveilCD} (\citeyear{Fu2024UnveilCD}), whose conclusion we directly use to prove our Theorem \ref{thm1}. Assumption 2 allows the covariate $Z$ to be unbounded,  provided that  its density function has sub-Gaussian tails.

\section{S2. Theoretical Results}
\subsection{Parameters of ReLU neural networks}
We use a ReLU neural network $\widehat{s}(x,z,t)\in \mathcal{F}$ to approximate the score function $\nabla \log p_{t}(\cdot|z)$. $\mathcal{F}$ is the following class of neural networks:
\begin{equation*}
    \begin{aligned}
        &\mathcal{F}(M_t,W,\kappa,L,K):=\Bigg\{ s(x,z,t)\\
        &~~~~~~=(A_L\sigma(\cdot)+b_L)\circ \cdots \circ (A_1(x,z^{\top},t)^{\top}+b_1):\\ 
        &~~~~~~A_i\in \mathbb{R}^{d_i\times d_{i+1}}, ~b_i\in \mathbb{R}^{d_{i+1}},~\max_i d_i\leq W,\\
        &~~~~~~\sup_{x,z}\|s(x,z,t)\|_{\infty}\leq M_t, \\ &~~~~~~\max_i\|A_i\|_{\infty}\vee \|b_i\|_{\infty}\leq \kappa,\\
        &~~~~~~\sum_{i=1}^L(\|A_i\|_0+\|b_i\|_0)\leq K \Bigg\},
    \end{aligned}
\end{equation*}
where $\sigma(\cdot)$ is the ReLU activation, $\|\cdot\|_{\infty}$ is the maximal magnitude of entries and $\|\cdot\|_0$ is the number of nonzero entries. The complexity of this network class is controlled by the number of layers $L$, the number of neurons of each layer $W$, the magnitude of the network parameters $\kappa$, the number of nonzero parameters $K$, and the magnitude of the neural network output $M_t$. According to \citeauthor{Fu2024UnveilCD} (\citeyear{Fu2024UnveilCD}), we  set $M_t=O(\sqrt{\log S/(1-\exp(-t))})$, $W=O(S\log ^7 S)$, $\kappa=\exp(O(\log ^4 S))$, $L=O(\log^4 S)$, and $K=O(S\log^9S)$. Furthermore, by choosing the network size parameter $S=N^{\frac{d_x+d_z}{d_x+d_z+2(k+\alpha)}}$, we can establish the convergence result stated in Theorem \ref{thm1}.

\subsection{Proof of Theorem \ref{thm1}}
\begin{proof}[Proof]
According to Lemma D.7 in \citeauthor{Fu2024UnveilCD} (\citeyear{Fu2024UnveilCD}), replacing their $\{s_i^{\prime}\}_{i=1}^{N}$ with our $\{X_i^{\text{U}}\}_{i=1}^{N}$ and their $\{s_i , a_i\}_{i=1}^{N}$ with our $\{Z_i^{\text{U}}\}_{i=1}^{N}$, there exists a constant $C_1 > 0$ such that

\begin{equation} \label{Bound of Rs}
\begin{aligned}
    &\mathbb{E}_{\{X_i^{\text{U}},Z_i^{\text{U}}\}_{i=1}^{N}}[\mathcal{R}(\widehat{s})]\\
    &\leq C_1 \left(\log{\frac{1}{t_{\min}}}\right) N^{- \frac{2(k+{\alpha})}{{d_x}+{d_z}+{2(k+{\alpha})}}} {(\log N)}^{\max(17, k+\alpha)},
\end{aligned}
\end{equation} where $\mathcal{R}(\widehat{s})$ is defined as

\begin{equation*}
\begin{array}{l}
    \mathcal{R}(\widehat{s})
    =\int_{t_{\min}}^{T}\frac{1}{T-{t}_{\min}}
    \mathbb{E}_{(X(t),Z)}\big{[} \| \widehat{s}(X(t),Z,t)
    \\ \\ ~~~~~~~~~~~~~~~~~~~~~~~~~~~~~~~~~~~~~~~~~~~~~~~-\nabla \log p_{t}(X(t)|Z) \|_{2}^{2}\big{]}dt. 

\end{array}
\end{equation*}

Then, according to the proof of Proposition 4.5 in  \citeauthor{Fu2024UnveilCD} (\citeyear{Fu2024UnveilCD}),
there exists positive constants ${C_2},{C_3},{C_4}$ such that

\begin{equation*}
\begin{array}{l}
    d_{\text{TV}}(p(\cdot|Z),\widehat{p}(\cdot|Z))
    \\ \\ \leq C_2 \sqrt{t_{\min}}(\log{\frac{1}{t_{\min}}})^{({d_x}+1) /2} + C_3 \exp(-T)
    \\ \\~~~~+C_4 \big{[}\int_{t_{\min}}^{T}\frac{1}{2}\int_{X(t)}
    p_t(X(t)|Z) \| \widehat{s}(X(t),Z,t)
    \\ \\ ~~~~
    ~~~~~~~~~~~~~~~~~~~~~~~~-\nabla \log p_{t}(X(t)|Z) \|^{2}dX(t)dt\big{]}^{\frac{1}{2}}.

\end{array}
\end{equation*}

Then taking expectation w.r.t. $Z$ and using Jensen's inequality and the definition of $\mathcal{R}(\widehat{s})$, we have

\begin{equation*}
\begin{array}{l}
    \mathbb{E}_{Z}\big{[}d_{\text{TV}(p(\cdot|Z),\widehat{p}(\cdot|Z))}\big{]}
    \\ \\ \leq C_2 \sqrt{t_{\min}}(\log{\frac{1}{t_{\min}}})^{({d_x}+1) /2} + C_3 \exp(-T)
    \\ \\~~~~+C_4 \mathbb{E}_{Z}\big{[}\int_{t_{\min}}^{T}\frac{1}{2}\int_{X(t)}
    p_t(X(t)|Z) \| \widehat{s}(X(t),Z,t)
    \\ \\ ~~~~
    ~~~~~~~~~~~~~~~~~~~~~~~~-\nabla \log p_{t}(X(t)|Z) \|^{2}dX(t)dt\big{]}^{\frac{1}{2}}
    \\ \\ \leq C_2 \sqrt{t_{\min}}(\log{\frac{1}{t_{\min}}})^{({d_x}+1) /2} + C_3 \exp(-T)
    \\ \\ ~~~~+C_4 \sqrt{\frac{T}{2}{\mathcal{R}(\widehat{s})}}.

\end{array}
\end{equation*}

Here we set $T = \frac{2(k+{\alpha})}{{d_x}+{d_z}+2(k+{\alpha})}\log{N}$ and take expectation w.r.t. 
$\{X_i^{\text{U}},Z_i^{\text{U}}\}_{i=1}^{N}$, and still by Jensen's inequality we have that there exists a constant $C_5>0$ such that 

\begin{equation*}
\begin{array}{l}
    \mathbb{E}_{\{X_i^{\text{U}},Z_i^{\text{U}}\}_{i=1}^{N}}
    \big{[}\mathbb{E}_{Z}\big{[}d_{\text{TV}(p(\cdot|Z),\widehat{p}(\cdot|Z))}\big{]}\big{]}
    \\ \\ \leq C_2 \sqrt{t_{\min}}(\log{\frac{1}{t_{\min}}})^{({d_x}+1) /2} + C_3 N^{-\frac{2(k+{\alpha})}{{d_x}+{d_z}+2(k+{\alpha})}}
    \\ \\~~~~ +C_5 \sqrt{\log N}~
    \mathbb{E}_{\{X_i^{\text{U}},Z_i^{\text{U}}\}_{i=1}^{N}}
    \left[\sqrt{{\mathcal{R}(\widehat{s})}}\right]
    \\ \\ \leq C_2 \sqrt{t_{\min}}(\log{\frac{1}{t_{\min}}})^{({d_x}+1) /2} + C_3 N^{-\frac{2(k+{\alpha})}{{d_x}+{d_z}+2(k+{\alpha})}}
    \\ \\~~~~ +C_5 \sqrt{\log N}
    \sqrt{\mathbb{E}_{\{X_i^{\text{U}},Z_i^{\text{U}}\}_{i=1}^{N}}
    \big{[}{\mathcal{R}(\widehat{s})}\big{]}}.

\end{array}
\end{equation*}

Then we can use the bound of $\mathbb{E}_{\{X_i^{\text{U}},Z_i^{\text{U}}\}_{i=1}^{N}}[\mathcal{R}(\widehat{s})]$ in (\ref{Bound of Rs}) to get

\begin{equation*}
\begin{array}{l}
    \mathbb{E}_{\{X_i^{\text{U}},Z_i^{\text{U}}\}_{i=1}^{N}}
    \big{[}\mathbb{E}_{Z}\big{[}d_{\text{TV}(p(\cdot|Z),\widehat{p}(\cdot|Z))}\big{]}\big{]}
    \\ \\ \leq C_2 \sqrt{t_{\min}}(\log{\frac{1}{t_{\min}}})^{({d_x}+1) /2} + C_3 N^{-\frac{2(k+{\alpha})}{{d_x}+{d_z}+2(k+{\alpha})}}
    \\ \\~~~~ + \sqrt{C_1} C_5 \sqrt{\log{\frac{1}{t_{\min}}}} N^{-\frac{(k+{\alpha})}{{d_x}+{d_z}+{2(k+{\alpha})}}} (\log N)^{\max \left(9,\frac{k+\alpha + 1}{2}\right)}.

\end{array}
\end{equation*}

Here we take $t_{\min}=N^{-\frac{4(k+{\alpha})}{{d_x}+{d_z}+{2(k+{\alpha})}}-1}$ such that
$\sqrt{t_{\min}}(\log{\frac{1}{t_{\min}}})^{({d_x}+1) /2} \leq N^{-\frac{2(k+{\alpha})}{{d_x}+{d_z}+2(k+{\alpha})}}$ for sufficiently large $N$, and then we have

\begin{equation*}
\begin{array}{l}
    \mathbb{E}_{\{X_i^{\text{U}},Z_i^{\text{U}}\}_{i=1}^{N}}\mathbb{E}_Z[d_{\text{TV}}(p(\cdot|Z),\widehat{p}(\cdot|Z))] \\ \\
    =O\left( N^{-\frac{k+{\alpha}}{{d_x}+{d_z}+2(k+{\alpha})}} \cdot \left(\log N \right) ^{\max(\frac{19}{2},\frac{k+\alpha + 2}{2})} \right). 
\end{array}
\end{equation*}
\end{proof}

\subsection{Proof of Theorem \ref{thm2}}
\begin{proof}[Proof]
Let $\acute{\bm{X}}$ be an additional copy drawn from $\widehat{p}^{(n)}(\cdot|\bm{Z})$, independent of $\bm{Y}$ and also of $\bm{X},\widehat{\bm{X}}^{(1)},\ldots,\widehat{\bm{X}}^{(B)}$. 
For any test statistic $\bm{T}$, define $\mathcal{A}_{\alpha}^B:=\bigg{\{}(\bm{x},\bm{x}^{(1)},\ldots, \bm{x}^{(B)})\Big|~\big{[}1+\sum_{b=1}^B\bm{\mbox{I}}\{\bm{T}(\bm{x}^{(b)},\bm{Y},\bm{Z})\geq \bm{T}(\bm{x},\bm{Y},\bm{Z})\}\big{]}\big/(1+B)\leq \alpha \bigg{\}}$. Note that in our case, the statistic $\bm{T}$ is selected to be $\widehat{\mbox{CMI}}$. Then, it follows that
    \begin{align*}
    P(p\leq \alpha|\bm{Y},\bm{Z})&=P((\bm{X},\widehat{\bm{X}}^{(1)},\ldots ,\widehat{\bm{X}}^{(B)})\in \mathcal{A}_{\alpha}^B|\bm{Y},\bm{Z}) \\
    &=P((\acute{\bm{X}},\widehat{\bm{X}}^{(1)},\ldots ,\widehat{\bm{X}}^{(B)})\in \mathcal{A}_{\alpha}^B|\bm{Y},\bm{Z}) \\
    &~~~~+P((\bm{X},\widehat{\bm{X}}^{(1)},\ldots ,\widehat{\bm{X}}^{(B)})\in \mathcal{A}_{\alpha}^B|\bm{Y},\bm{Z}) \\
    &~~~~-P((\acute{\bm{X}},\widehat{\bm{X}}^{(1)},\ldots ,\widehat{\bm{X}}^{(B)})\in \mathcal{A}_{\alpha}^B|\bm{Y},\bm{Z}) \\
    &\leq P((\acute{\bm{X}},\widehat{\bm{X}}^{(1)},\ldots ,\widehat{\bm{X}}^{(B)})\in \mathcal{A}_{\alpha}^B|\bm{Y},\bm{Z}) \\
    &~~~~+d_{\text{TV}}\{(\bm{X},\widehat{\bm{X}}^{(1)},\ldots ,\widehat{\bm{X}}^{(B)}|\bm{Y},\bm{Z}), \\
    &~~~~~~~~~~~~~~~~~~~~~~(\acute{\bm{X}},\widehat{\bm{X}}^{(1)},\ldots ,\widehat{\bm{X}}^{(B)}|\bm{Y},\bm{Z}) \} \\
    &=P((\acute{\bm{X}},\widehat{\bm{X}}^{(1)},\ldots ,\widehat{\bm{X}}^{(B)})\in \mathcal{A}_{\alpha}^B|\bm{Y},\bm{Z}) \\
    &~~~~+d_{\text{TV}}(p^{(n)}(\cdot|\bm{Z}),\widehat{p}^{(n)}(\cdot|\bm{Z}) ),
    \end{align*} where the last equality follows from the property of TV distance and the fact that $\bm{X},\acute{\bm{X}},\widehat{\bm{X}}^{(1)},\ldots ,\widehat{\bm{X}}^{(B)}$ are independent conditionally on $\bm{Y}$ and $\bm{Z}$. It is clear that $\acute{\bm{X}},\widehat{\bm{X}}^{(1)},\ldots ,\widehat{\bm{X}}^{(B)}$ are i.i.d. conditional on $\bm{Y}$ and $\bm{Z}$, and are thus exchangeable. By the definition of $\mathcal{A}_{\alpha}^B$ and the property of rank test \cite{kuchibhotla2020exchangeability}, we have $P((\acute{\bm{X}},\widehat{\bm{X}}^{(1)},\ldots ,\widehat{\bm{X}}^{(B)})\in \mathcal{A}_{\alpha}^B|\bm{Y},\bm{Z})\leq \alpha$. Finally, we have $P(p\leq \alpha|\bm{Y},\bm{Z})\leq \alpha+d_{\text{TV}}(p^{(n)}(\cdot|\bm{Z}),\widehat{p}^{(n)}(\cdot|\bm{Z}) )$.
\end{proof}

\subsection{Proof of Corollary \ref{corollary}}
\begin{proof}[Proof]
According to the argument under Theorem \ref{thm2}, we only need to show $\mathbb{E}[d_{\text{TV}}(p^{(n)}(\cdot|\bm{Z}),\widehat{p}^{(n)}(\cdot|\bm{Z}))]=o(1)$. By the definition of TV distance, we have
\begin{align*}
    &d_{\text{TV}}(p^{(n)}(\cdot|\bm{Z}),\widehat{p}^{(n)}(\cdot|\bm{Z})) \\
    &=\frac{1}{2}\int \cdots \int \left |\prod_{i=1}^np(x_i|Z_i)-\prod_{i=1}^n\widehat{p}(x_i|Z_i)\right |dx_1\cdots dx_n \\
    &\leq \frac{1}{2}\int \cdots \int \left |p(x_1|Z_1)-\widehat{p}(x_1|Z_1)\right|\cdot \\
    &~~~~~~~~~~~~~~~~~~~~~~~~~~~~~~~~~~~~~~~~~~~\prod_{j=2}^np(x_j|Z_j)dx_1\cdots dx_n \\
    &~~~~+\sum_{i=2}^{n-1}\bigg{(} \frac{1}{2}\int \cdots \int \prod_{j=1}^{i-1}\widehat{p}(x_j|Z_j)\cdot \left |p(x_i|Z_i)-\widehat{p}(x_i|Z_i)\right |\cdot \\
    &~~~~~~~~~~~~~~~~~~~~~~~~~~~~~~~~~~~~~~~~~\prod_{j=i+1}^{n}p(x_j|Z_j)dx_1\cdots dx_n \bigg{)} \\
    &~~~~+\frac{1}{2}\int \cdots \int  \prod_{j=1}^{n-1}\widehat{p}(x_j|Z_j)\cdot \left|p(x_n|Z_n)-\widehat{p}(x_n|Z_n)\right | \\
    &~~~~~~~~~~~~~~~~~~~~~~~~~~~~~~~~~~~~~~~~~~~~~dx_1\cdots dx_n \\
    &=\sum_{i=1}^n\left ( \frac{1}{2}\int |p(x_i|Z_i)-\widehat{p}(x_i|Z_i)|dx_i \right ) \\
    &=\sum_{i=1}^n d_{\text{TV}}(p(\cdot|Z_i),\widehat{p}(\cdot|Z_i)),
\end{align*} which implies that
\begin{align*}
    &\mathbb{E}[d_{\text{TV}}(p^{(n)}(\cdot|\bm{Z}),\widehat{p}^{(n)}(\cdot|\bm{Z})] \\
    &\leq n \mathbb{E}[d_{\text{TV}}(p(\cdot|Z),\widehat{p}(\cdot|Z))] \\
    &=O\left( n\cdot N^ {-\Gamma_1(k,\alpha)} \cdot \left(\log N \right) ^{\Gamma_2(k,\alpha)} \right) \\
    &=o(1),
\end{align*} where the first equality follows from Theorem \ref{thm1} and the last equality follows from the assumption on the sample size.
\end{proof} 

\begin{figure}[ht]
    \centering
    \begin{minipage}[t]{0.83\linewidth}
        \centering
        \includegraphics[width=1\linewidth]{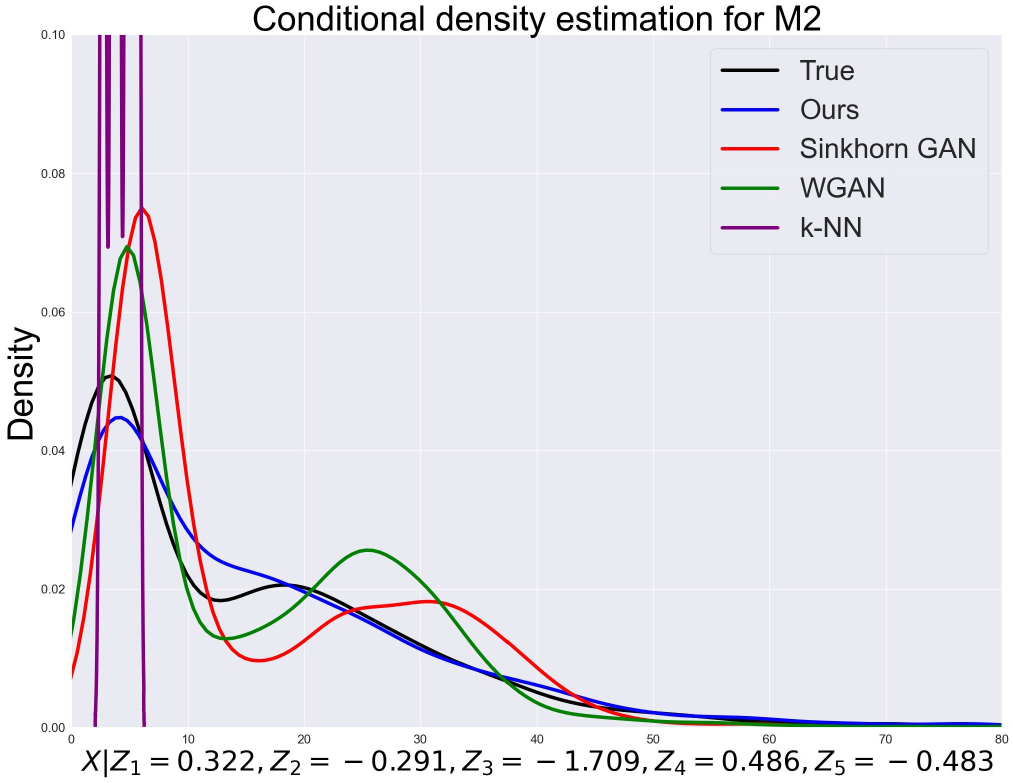}
    \end{minipage}
    \begin{minipage}[t]{0.83\linewidth}
        \centering
        \includegraphics[width=1\linewidth]{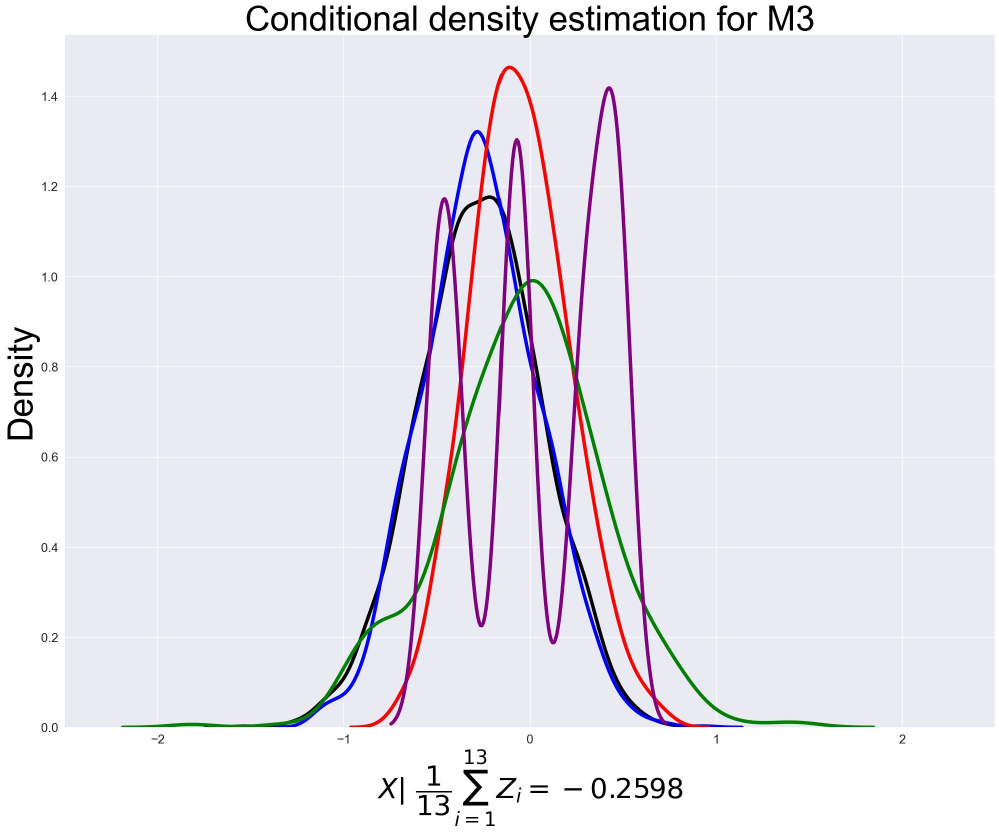}
    \end{minipage}
    \caption{Comparisons of conditional density estimators on Model M2 and M3. 
    }
    \label{fig:M2 to M3}
 \end{figure}


\section{S3. Additional Results on Synthetic Data}
\subsection{Scenario 1: $X$ and $Y$ are univariate}

\begin{figure*}[htb!]
    \centering
    \begin{minipage}{1\linewidth}
        \centering
        \includegraphics[width=0.85\linewidth]{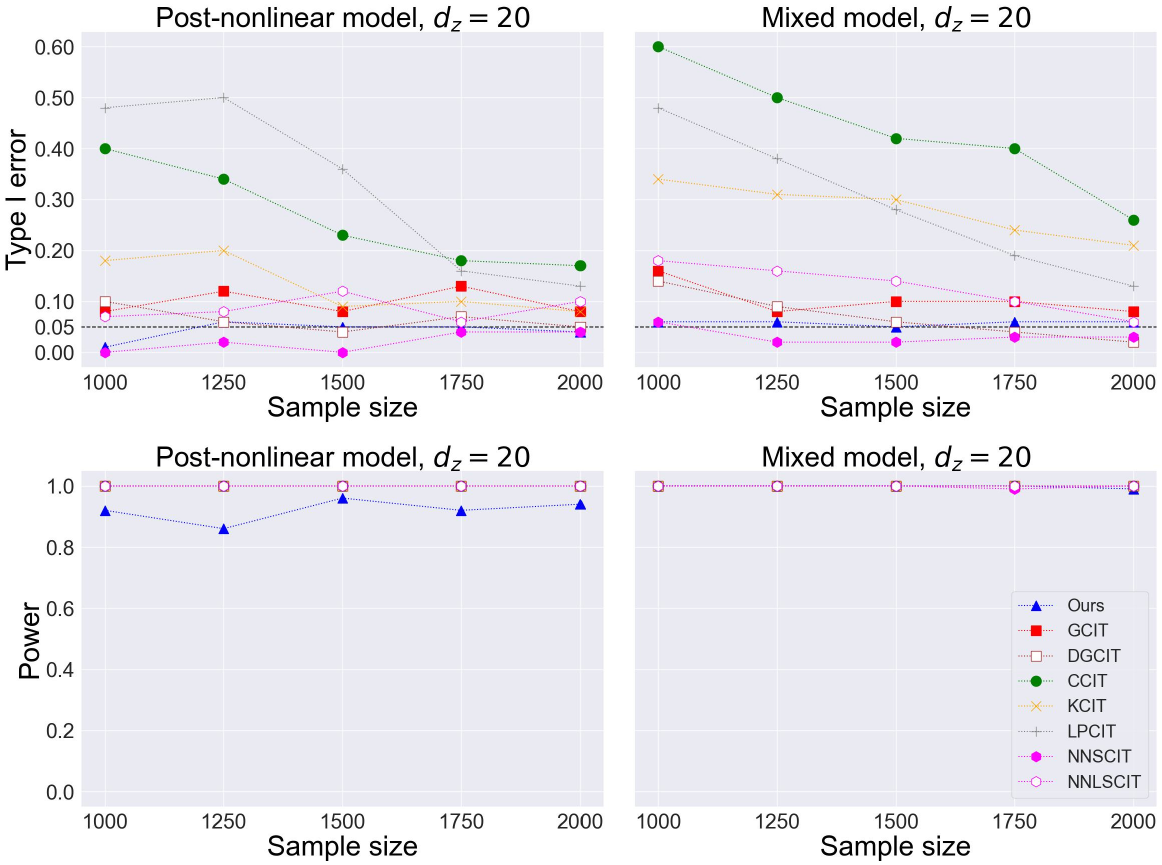}
    \end{minipage}
    
    \caption{Performance of various methods in terms of type I error and power under $H_1$  on the post-nonlinear model (\ref{fork_structue}) and mixed model (\ref{dis_and_continuous}) across different sample sizes with $d_z=20$. For our method, we set $n=500$ and vary $N$  from 500 to 1500.}
    \label{simulation_fork_dz20}
 \end{figure*}

We perform additional simulation studies using the synthetic dataset generated as detailed in Scenarios I and II of the main paper. We vary the sample size from $1000$ to $2000$ while keeping the dimension of $Z$ fixed at $20$. For our CDCIT method, $N$ ranges from 500 to 1500 to train the conditional sampler, and $n=500$ is fixed to compute the test statistic. The results are shown in Figure \ref{simulation_fork_dz20}. It can be observed that in both the post-nonlinear and mixed models, our test and NNSCIT consistently achieve good and stable performance in terms of type I error and power under $H_1$. On the other hand, LPCIT, CCIT, and KCIT exhibit high power under $H_1$ but exhibit significantly large  type I errors. Although DGCIT, GCIT, and NNLSCIT demonstrate adequate power under $H_1$, they also inflate type I errors in some scenarios, particularly under the mixed model.

\begin{figure*}[htb!]
    \centering
    \begin{minipage}{0.8\linewidth}
        \centering
        \includegraphics[width=1\linewidth]{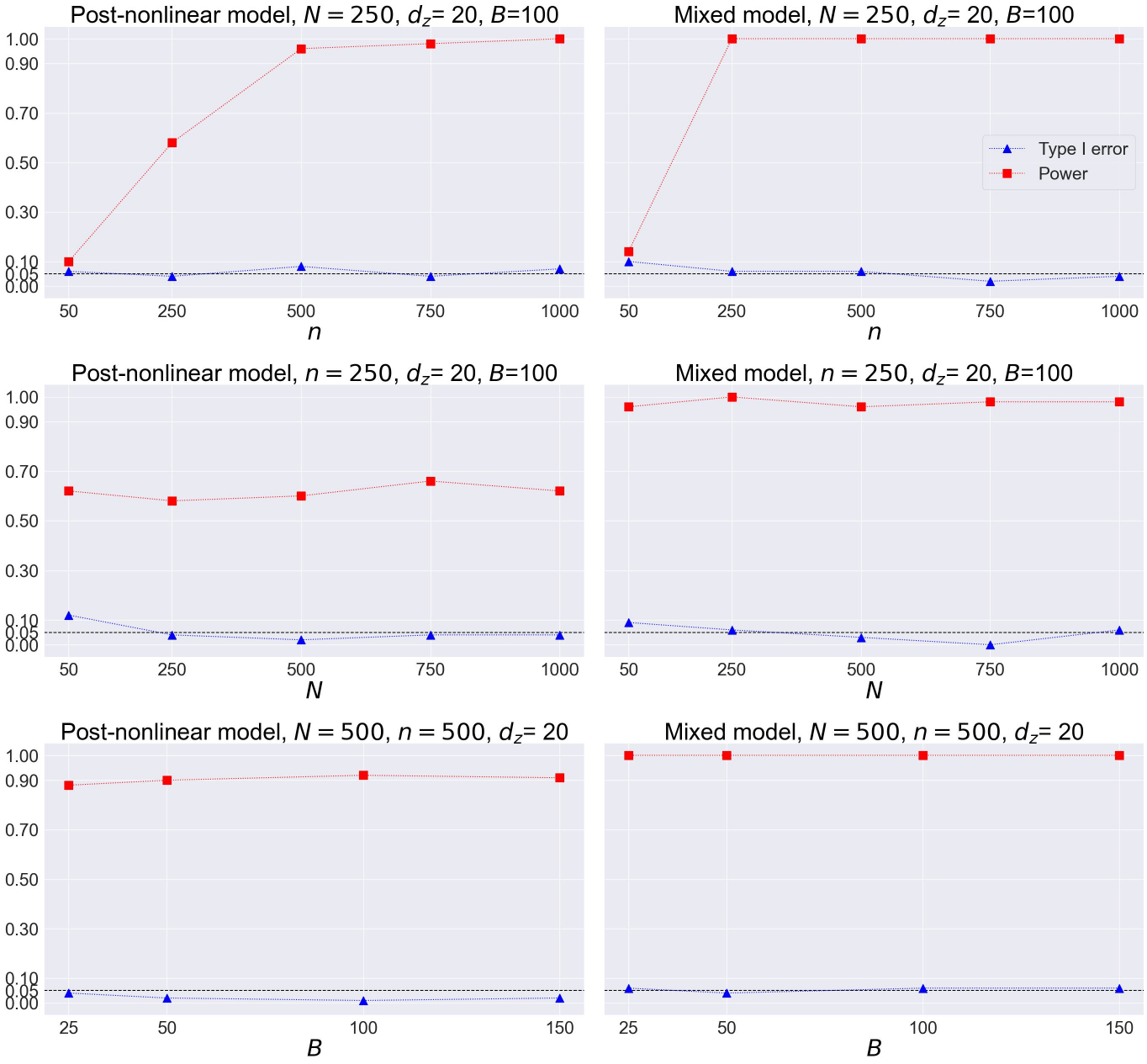}
    \end{minipage}
    
    \caption{CDCIT's performance in terms of type I error and power under $H_1$  on the post-nonlinear models (\ref{fork_structue}) and mixed model  (\ref{dis_and_continuous}) with $d_z=20$,  across different $(N, n, B)$ configurations.}
    \label{different_N_n_B}
 \end{figure*}

Furthermore, we investigate the performance of the proposed CDCIT method across various combinations of $N$ and $n$, and examine the impact of the number of repetitions $B$ in Algorithm \ref{main:algorithm} on CDCIT's performance. Specifically, we first fix $N=250$ and vary $n$ from 50 to 1000; then, we fix $n=250$ and vary $N$ from 50 to 1000; finally, we vary $B$ from 25 to 150.  The results are presented in Figure \ref{different_N_n_B}. It is shown  that $n$ plays a critical role in determining the power of CDCIT. The power increases rapidly with the growth of $n$ before eventually plateauing. However, $n$ does not significantly affect the type I error. We also observe that a smaller $N$ leads to inflated type I error, likely due to challenges in training the conditional diffusion models. Additionally, $N$ has no noticeable effect on power. Finally, we observe that the influence of the number of repetitions $B$ on the proposed method is minimal, indicating that our CDCIT is not sensitive to this parameter. 

\subsection{Scenario 2: $X$ and $Y$ are multivariate}
Our method can be extended to scenarios where $X$ and $Y$ are multivariate. To verify the effectiveness in such cases, we define $(X, Y, Z)$ under $H_0$ and $H_1$ as follows: 
\begin{eqnarray*}
    && H_0:X=f(A_{x}^TZ+0.33\epsilon_{x}),~Y=g(A_{y}^TZ+0.33\epsilon_{y}), \nonumber \\
    && H_1:X=f(A_{x}^TZ)+0.33\epsilon_{x},~Y=h(A_{y}^TZ)+0.33\epsilon_{x}.
\end{eqnarray*}
The entries of $A_x \in \mathcal{R}^{d_z \times d_x}$ and $A_y \in \mathcal{R}^{d_z \times d_y}$ are randomly and uniformly sampled from $[0,1]$ and then normalized to the unit $l_1$ norm. The noise variables $\epsilon_x \in \mathcal{R}^{d_x}$ and $\epsilon_y  \in \mathcal{R}^{d_y}$  are independently sampled from normal distributions with mean zero and identity matrix. For simplicity, we set $d_x=d_y$. We set $f$, $g$, and $h$ to be randomly sampled from $\left\{x, x^{2}, x^{3}, \mbox{tanh}(x), \cos(x)\right\}$ and $Z\sim N(0,1)$ under both $H_0$ and $H_1$. The dimensions of $X$, $Y$, and $Z$ are set to (5, 5, 10), (10, 10, 5), (50, 50, 100) and (100, 100, 50), respectively. We use $N=500$ to train the conditional sampler and $n=500$ to compute the test statistic in our test. The results are provided in Table \ref{multiva_xy}. We observe that, our CDCIT controls type I error very well and achieves high power under $H_1$ in scenarios with multivariate $X$ and $Y$..

\begin{table*}[ht]
\centering
\begin{tabular}{c|cc}
\hline
  $(d_x,d_y,d_z)$   &  Type I error & Power\\
  \hline
  (5, 5, 10)   &  0.05 & 1 \\
  (10, 10, 5) & 0.03 & 0.96 \\
  (50,50,100) & 0.03 & 1 \\
  (100,100,50) & 0.06 & 1 \\
  \hline
\end{tabular}
\caption{The empirical results of our test in the scenario where $X$ and $Y$ are multivariate.}
\label{multiva_xy}
\end{table*}

\section{S4. Real Data Analysis}
\subsection{Study 1: Real breast cancer dataset}
\begin{table*}[ht]
\centering
\begin{tabular}{c|cccccccc}
\hline
Method &Ours &GCIT & DGCIT & CCIT & KCIT & LPCIT & NNSCIT & NNLSCIT\\
\hline
Number of selected genes & 8 & 29 & 97 & 43 & 34 & 31 & 24 & 9 \\
\hline
Average MSE & 0.696 & 0.706 & 0.742 & 0.773 & 0.733 & 0.718 & 0.724 & 0.737 \\
\hline
Standard deviation & 0.010 & 0.011 & 0.015 & 0.028 & 0.024 & 0.014 & 0.008 & 0.013 \\
\hline
\end{tabular}
\caption{Number of genes selected by each method and the average prediction MSE for tumor size using these genes along with clinical covariates.}
\label{use_gene_predict_tumor_size}
\end{table*}

We conduct a comparative evaluation of our method  against SOTA methods using a breast cancer genomic dataset from \citeauthor{nature_breast_cancer} \shortcite{nature_breast_cancer}. The data can be downloaded from \url{https://www.cbioportal.org/study/summary?id=brca_metabric}. Following \citeauthor{liu2022fast} \shortcite{liu2022fast}, we focus on the 1077 patients with estrogen-receptor-positive (ER+) status. For each patient, data includes 169 genes and the 5 discrete clinical covariates: \textit{HER2} status,  receipt of hormone therapy,  menopausal status, survival status after a follow-up period, and receipt of  radiation therapy.

We aim to identify the genes that influence tumor size, conditional on the remaining genes and the discrete clinical covariates, while controlling the false discovery rate (FDR) at level $0.2$ \cite{efron2004selection}. These conditionally dependent (CD) relationships can provide crucial evidence for clinicians to identify target genes for cancer and develop targeted therapies \cite{APC_breast_cancer}. Specifically,  we  select each of the 169 genes as $X$, set tumor size as $Y$, designate the remaining 168 genes and 5 clinical covariates as $Z$, and  then  employ various CI testing methods to test the CD. 
We conduct multiple testing using AdaptZ~\citep{sun2007oracle}, which under certain conditions is optimal in minimizing the false nondiscovery rate while controlling FDR.  Our CDCIT identifies 8 genes as  conditionally dependent on  tumor size. The numbers of genes identified by other competing methods are shown in Table \ref{use_gene_predict_tumor_size}.

We use the genes selected by each method along with clinical covariates to predict tumor size using XGBoost \cite{chen2016xgboost}. We evaluate the predictions using the mean squared error (MSE) as the metric; a smaller MSE indicates that the selected genes are more likely to contribute to tumor size. 
We repeat the XGBoost procedure  50 times and present the average MSE in Table \ref{use_gene_predict_tumor_size} for each method. It is observed that our method achieves the lowest MSE. 
We provide additional real data analysis results in Supplementary Materials.

Figure \ref{realdata_venn} presents a Venn diagram illustrating the  conditionally dependent (CD) genes identified by  seven  algorithms: GCIT, CCIT, KCIT, LPCIT, NNSCIT, NNLSCIT, and our method.  Our algorithm identifies 8 genes as  conditionally dependent on tumor size, most of which overlap with genes identified by at least one other algorithm. 
Note that DGCIT is excluded from the diagram as it identifies an excessively large number of genes as CD, approximately 57\%. 

We now turn our attention to the more specific information of the 8 CD genes identified by our algorithm. As detailed in Table \ref{breast_cancer_22_genes}, 6 out of these 8 genes were classified as CD by two or more other CIT methods, with the exceptions of \textit{MLLT4} and \textit{NR3C1}. 
It is noted that \textit{DNAH11} has been found to be significantly associated with breast cancer \cite{DNAH11_breast_cancer}. \textit{FOXP1} regulates estrogen signaling to drive breast cancer cell growth and is linked to better outcomes in patients undergoing tamoxifen therapy \cite{FOXP1_breast_cancer}. \textit{FOXO3} acts as a tumor suppressor in breast cancer and is notably affected by widely used anti-breast cancer drugs, including paclitaxel, simvastatin, and gefitinib \cite{FOXO3_breast_cancer}. \textit{MLLT4} has been identified as a suppressor gene for mutation-driven tumors \cite{MLLT4_breast_cancer}. \textit{NR3C1} plays a role in tumor cell differentiation and proliferation \cite{NR3C1_breast_cancer}. The deletions of \textit{PPP2R2A} are correlated to a subgroup of breast cancer exhibiting poor survival \cite{PPP2R2A_breast_cancer}. Mutations of \textit{SF3B1} cause changes in RNA splicing and are tied to certain types of breast cancer, making \textit{SF3B1} a possible treatment target \cite{SF3B1_breast_cancer}. Additionally, the mutation of \textit{TP53} is prevalent in breast cancer, accounting for up to 40\% of cases \cite{TP53_breast_cancer}. 


Due to the absence of ground truth, we use genes selected by various CI testing methods along with clinical information, to estimate tumor size, thereby providing  further evaluations of these CI testing approaches. The prediction mean squared error (MSE) is used as a metric to assess the performance of each CI testing method. For a given CI testing, a smaller MSE indicates that the selected genes are more likely to contribute to tumor formation. The MSE results are presented in Table \ref{use_gene_predict_tumor_size}, with the detailed calculation process described in Supplementary Section S6.

\begin{table*}[ht]
\centering
\begin{tabular}{ccccccccc}
\hline
Genes & Ours &GCIT & DGCIT & CCIT & KCIT & LPCIT & NNSCIT & NNLSCIT\\
\hline
\textit{DNAH11  }& CD & CD & CI & CI & CI & CD & CI & CI \\
\textit{FOXP1   }& CD & CD & CI & CI & CD & CD & CI & CI \\
\textit{FOXO3   }& CD & CD & CI & CI & CI & CD & CD & CI \\
\textit{MLLT4   }& CD & CI & CD & CI & CI & CI & CI & CI \\
\textit{NR3C1   }& CD & CI & CD & CI & CI & CI & CI & CI \\
\textit{PPP2R2A }& CD & CD & CD & CI & CD & CD & CI & CD \\
\textit{SF3B1   }& CD & CD & CD & CD & CD & CD & CI & CI \\
\textit{TP53    }& CD & CI & CD & CI & CD & CD & CI & CD \\
\hline

\end{tabular}
\caption{Results of various methods on the 8 genes identified as conditionally dependent (CD) on tumor size by our CDCIT.}
\label{breast_cancer_22_genes}
\end{table*}

\begin{figure}[htb!]
\centering
\includegraphics[width=0.45\textwidth]{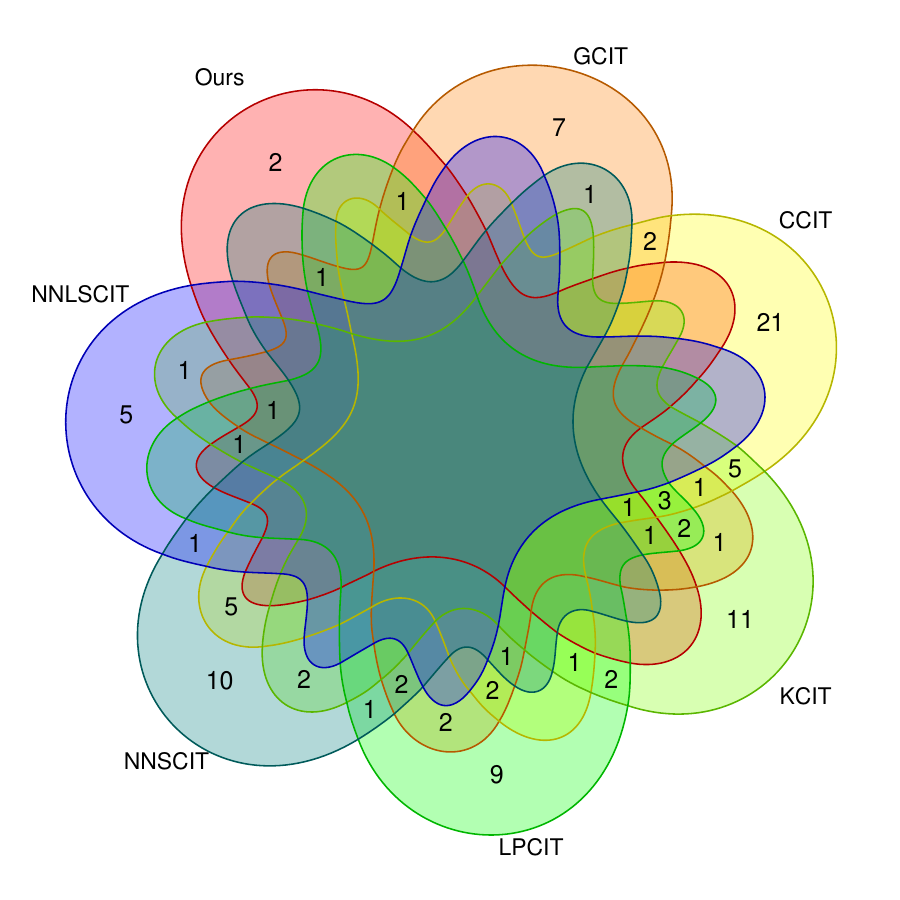}
\caption{Venn diagram of CD genes detected by 7 algorithms: GCIT, CCIT, KCIT, LPCIT, NNSCIT, NNLSCIT and our method.} 
\label{realdata_venn}
\end{figure}

\subsection{Study 2: Real Flow-Cytometry dataset}
The Flow-Cytometry dataset is a widely used benchmark in the field of causal structure learning \cite{ng2020role, zhucausal}. This dataset captures the expression levels of proteins and phospholipids in human cells \cite{sachs2005causal}. The data can be obtained from the website https://www.science.org/doi/10.1126/science.1105809. In our evaluation, we consider the consensus graph proposed in \citeauthor{mooij2013cyclic} \shortcite{mooij2013cyclic} as the ground truth, which has also been adopted by \citeauthor{sen2017model} \shortcite{sen2017model} and \citet{li2024k} for verifying CI relations. 

We specifically select 50 CI relations and 40 non-CI relations from this directed acyclic graph (DAG). The number of samples is 1755 and the dimension of $Z$ varies from 1 to 9. The rule used is that a node $X$ is independent of all other nodes $Y$ in a DAG when conditioned on its parents, children, and parents of children \cite{koller2009probabilistic}. Additionally, if there exists a direct edge between node $X$ and node $Y$ in a DAG, they are never conditionally independent given any other set of variables. As a result, the conditioning set $Z$ can be arbitrarily selected from the remaining nodes. 

To evaluate the performance of various tests, we utilize precision, recall, and F-score as evaluation metrics. Precision is calculated as TP/(TP+FP), where TP represents the number of true CI instances correctly identified, and FP represents the number of non-CI instances incorrectly identified as CI. Recall is calculated as TP/(TP+FN), where FN represents the number of CI instances not identified. The F-score is then computed as the harmonic mean of precision and recall, given by 2 $\times$ precision $\times$ recall / (precision + recall) \cite{li2024k}. 

Table \ref{Flow-Cytometry dataset} presents the results for all methods. Our tests achieves the highest recall and F-score, outperforming other approaches.

\begin{table}[ht]
\centering
\begin{tabular}{cccc}
\hline
CITs &Precision &Recall &F-score \\
\hline
Ours & 0.72 & 0.98 & 0.83 \\
GCIT & 0.74 & 0.80 & 0.77 \\
DGCIT & 0.95 & 0.44 & 0.60 \\
CCIT & 0.76 & 0.64 & 0.70 \\
KCIT & 0.70 & 0.62 & 0.66 \\
LPCIT & 0.68 & 0.74 & 0.71  \\
NNSCIT & 0.67 & 0.80 & 0.73 \\
NNLSCIT & 0.73 & 0.92 & 0.81 \\
\hline
\end{tabular}
\caption{CIT's performance on Flow-Cytometry dataset.}
\label{Flow-Cytometry dataset}
\end{table}

\section{S5. Computational Efficiency Analysis}

In this section, we compare the computational speeds of the considered methods across different dimensions $d_z$ of $Z$  and varying sample sizes. All experiments were conducted on a machine equipped with an 8-core i7-11800H 2.3GHz CPU and an NVIDIA RTX 3070 (Laptop) GPU.

\begin{figure}[htbp]
    \centering
    \begin{minipage}{1\linewidth}
        \centering
        \includegraphics[width=1\linewidth]{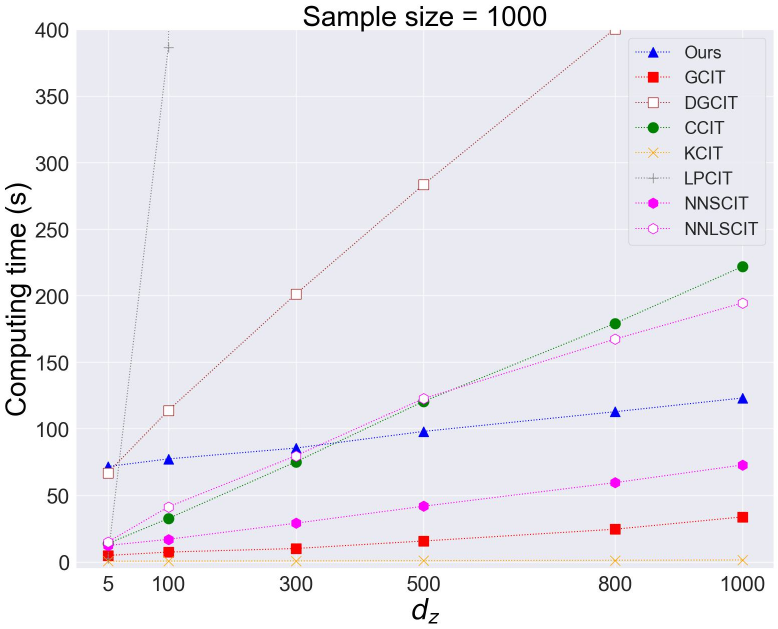}
    \end{minipage}

    \begin{minipage}{1\linewidth}
        \centering
        \includegraphics[width=1\linewidth]{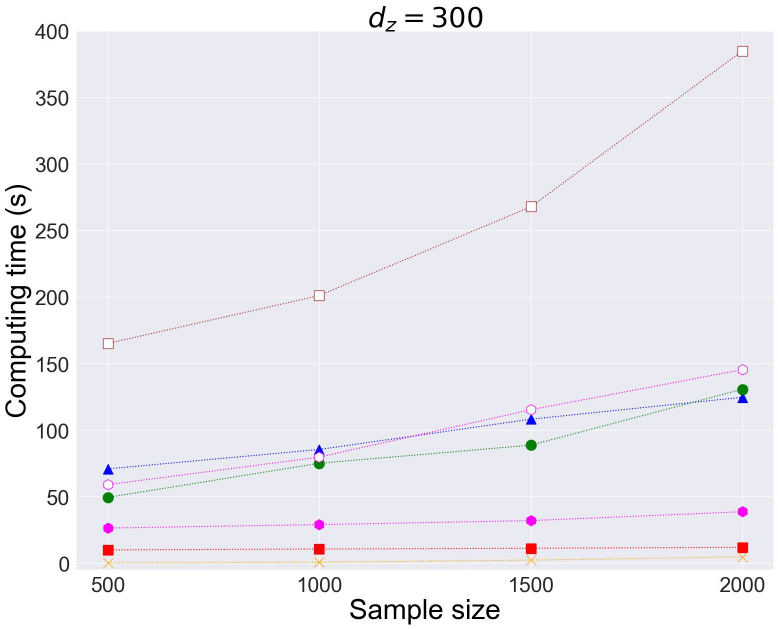}
    \end{minipage}
    
    \caption{Average running times in seconds as a function of the dimension of $Z$ ($d_z$) or sample size on the post-nonlinear model. The top panel displays the results for varying $d_z$, while the bottom panel shows the results for varying sample sizes. Due to the excessively long computing time of LPCIT, its results are not included in the bottom figure.}
    \label{computing_time_second}
 \end{figure}

Figure \ref{computing_time_second} presents the average computational time of all methods for 100 tests under the post-nonlinear model. Our test demonstrates high computational efficiency, even when handling high-dimensional conditioning sets and large sample sizes. In contrast, LPCIT and DGCIT are impractical for high-dimensional $Z$ due to their prohibitively long execution times.


\section{S6. Experiment Setup Details}
\label{Experiment_Details}

We use the Adam optimizer to train the deep neural network $\widehat{s}$, setting the learning rate to 0.01 and the number of epochs to 1500. In line with \citeauthor{ho2020denoising} \shortcite{ho2020denoising} and \citeauthor{han2022card} \shortcite{han2022card}, we set the maximum time index $T$ to 10, the minimum time index $t_{\text{min}}$ to 0.01, and the number of sampling steps $K$ to 1000.



When conducting experiments and calculating the MSEs of the quantiles of conditional distributions, as presented in Table \ref{quantile mse}, $\widehat{s}$ is composed of 3 hidden layers with 16 neurons, given that $d_z$ in M1-M3 is relatively low. The discriminators and generators of WGANs and Sinkhorn GANs share the same structure as $\widehat{s}$. The number of neighbors in k-NN is set to 7, following \citet{li2024k}. For conducting CRTs in the post-nonlinear and mixed models in the synthetic data analysis, $\widehat{s}$ is adjusted to have 3 hidden layers with 128 neurons, to accommodate potentially larger numbers of $d_z$. When testing the conditional dependence between genes and tumor size in the real data analysis, $\widehat{s}$ 
includes 3 Res-blocks with a width of 128. 


For predicting tumor size using selected genes and clinical information, we employ XGBoost as our regression model. Initially, we randomly split the dataset into a training set and a testing set with a 4:1 ratio. We then perform 5-fold cross-validation and grid search on the training set to identify the optimal parameters. The parameters tuned for XGBoost include the depth of the trees, learning rate, number of trees, sub-sample ratio, and the feature proportion used in constructing each tree. After determining the optimal parameters, we evaluate the MSE for tumor size prediction on the testing set. This procedure is repeated 50 times, and the results are summarized as the average MSEs in Table \ref{use_gene_predict_tumor_size}.

\section{S7. Additional Algorithms} \label{ccmial}

The 1-Nearest-Neighbor sampling algorithm and the  classifier-based CMI estimating procedure are presented in Algorithms \ref{NNS} and \ref{main:algorithm2}, respectively. 

\begin{algorithm}[htbp]
\caption{1-Nearest-Neighbor Sampling ({\bf 1-NN($V_1,V_2,n$)})}
\label{NNS}
\textbf{Input}: Datasets $V_1$ and $V_2$, both with sample size $n$ and $V=V_1\cup V_2$ consisting of $2n$ 
independently and identically distributed
(i.i.d.) samples from $p_{X,Y,Z}(x,y,z)$.\\
\textbf{Output}: A new data set $V'$ consists of $n$ samples.
\begin{algorithmic}[1] 
\STATE Let $V'=\emptyset$.
\FOR{$(X,Y,Z)$ in $V_2$}
\STATE Go to $V_1$ to find the sample $({X}', {Y}', {Z}')$ such that  ${Z}'$ is  the 1-nearest neighbor of $Z$   in terms of the $l_2$ norm
 \STATE  $V'=V'\cup \{ (X, {Y}', Z) \}$.
\ENDFOR
\STATE \textbf{return} $V'$
\end{algorithmic}
\end{algorithm}

\begin{algorithm}[htbp]
\caption{Classifier-based CMI Estimator}
\label{main:algorithm2}
\textbf{Input}: Dataset $V$ containing $2n$ i.i.d. samples drawn from $p_{X,Y,Z}(x,y,z)$.\\
\textbf{Output}: An estimator of CMI.
\begin{algorithmic}[1]
\STATE Equally split $V$ into two parts $V_1$ and $V_2$, each containing $n$ samples.
\STATE Apply Algorithm \ref{NNS} to generate a new dataset $V'$ with $|V'|=n$.
\STATE Form the labeled datasets $V_{f}=\{ (W_{i}^{f},l=1): {W_{i}^{f} \in V_2} \}$ and $V_{g}=\{ (W_{j}^{g},l=0): {W_{j}^{g} \in V'} \}$.
\STATE Divide $V_{f}$ into training and testing subsets $V_{f}^{\mbox{train}}$ and $V_{f}^{\mbox{test}}$, at a ratio of 2:1.
\STATE Similarly, split $V_{g}$ into training and testing subsets $V_{g}^{\mbox{train}}$ and $V_{g}^{\mbox{test}}$, at a ratio of 2:1.
\STATE Merge the datasets to form $V^{\mbox{train}}=V_{f}^{\mbox{train}}\cup V_{g}^{\mbox{train}}$ and $V^{\mbox{test}}=V_{f}^{\mbox{test}}\cup V_{g}^{\mbox{test}}$.
\STATE Train the classifier $C$ using $V^{\mbox{train}}$.
\STATE For each $w\in V_{0}^{test}$, where $V_{0}^{test}$ includes all features in $V^{\mbox{test}}$, compute the classifier-based predicted probability $P(l=1|w)$.
\STATE Calculate $\widehat{I}(X;Y|Z)$ as per formula (\ref{MIE}).
\STATE \textbf{return} $\widehat{I}(X;Y|Z)$.
\end{algorithmic}
\end{algorithm}

\section{S8. Further explanation of test statistic}

The critical step in transition from Equation (\ref{DVoptimal}) to Equation (\ref{MIE}) involves using the classifier’s predicted probability, $\alpha_m/(1-\alpha_m)$, to estimate the  likelihood ratio $f/g$. This approach is justified by Lemma 3 in \citeauthor{mukherjee2020ccmi} \shortcite{mukherjee2020ccmi}. Specifically speaking, the classifier is trained to minimize the binary-cross entropy (BCE) loss on the train set. The population BCE is defined as $$\text{BCE}(\alpha)=-(\mathbb{E}_{W,L}(L\log \alpha(W)+(1-L)\log (1-\alpha(W)))),$$ where the expectation is taken over the joint distribution of data and labels. Note that the samples from $f$ are labeled as $l=1$ and the samples from $g$ are labeled as $l=0$. Then the point-wise minimizer of BCE, $\alpha^*(w)$, is related to the likelihood ratio as $\frac{\alpha^*(w)}{1-\alpha^*(w)}=\frac{f}{g}=\frac{p(x,y,z)}{p(x, z)p(y|z)}$, where $\alpha^*(w)=P(L=1|W=w)$. This clarifies why we use the classifier to compute the predicted probability and how $\alpha_m/(1-\alpha_m)$ serves as an estimator for the likelihood ratio. Additionally, the $p(x,z)$ term appears in the likelihood ratio $f/g$ in Equation (\ref{DVoptimal}) and the likelihood ratio is transformed into the ratio of the classifier’s predicted probabilities.

When training the classifier, positive samples from $p(x,y,z)$ and negative samples from $p(x,z)p(y|z)$ are required. However, $p(y|z)$ is unknown, so we use the 1-NN method to model $p(y|z)$ to obtain negative samples, thus achieving data augmentation.

    

    

\end{document}